\documentclass[letter,12pt,TexShade,oneside]{book} 
\RequirePackage{fix-cm}
\usepackage{amssymb}
\usepackage{etex}
\usepackage{tikz}
\usepackage{tikzscale}
\usepackage{tcolorbox}
\usepackage{mdframed}
\usepackage{pgfplots}
\usepackage{epsfig}
\usepackage{etex}
\usepackage{amsmath}
\usepackage[landscape]{geometry}
\usepackage{lscape}
\usepackage[pdfa,colorlinks=true,urlcolor=black]{hyperref}
\usepackage{tabularx}
\usepackage{mathabx}
\usepackage{multirow}
\usepackage{moreverb}
\usepackage{multicol}
\usepackage{alltt}
\usepackage{setspace}
\usepackage{multirow}
\usepackage{lscape}
\usepackage[all]{xy}
\usepackage{color}
\usepackage{makeidx}
\usepackage{fancyvrb}
\usepackage{inputenc}
\usepackage{dsfont}
\usepackage{moreverb}
\usepackage{array}
\usepackage{supertabular}
\usepackage{hhline}
\usepackage{rotating}
\hypersetup{colorlinks=true, linkcolor=Brown, citecolor=Brown, filecolor=Brown, urlcolor=Brown}
\usepackage{epsfig}
\usepackage{calc}
\usepackage{listings}
\usepackage{mathtools}
\usepackage[T1]{fontenc}
\usepackage{bm}
\usepackage{theorem}
\usepackage{times,epsfig}
\usepackage{calc}
\usepackage{moresize}
\usepackage[caption=false]{subfig}
\usepackage{url}
\usepackage{multirow}
\usepackage[bf,rm,medium,compact]{titlesec}
\usepackage{times}  
\usepackage{pifont}     % to include symbols from different fonts (e.g.  \rtm)
\usepackage{eurosym}    % for the euro symbol, package by Henrik Theiling
\usepackage{graphicx}
\usepackage{booktabs}   % tables as required by journals
\usepackage{fancyhdr}                % fancy page headers/footers
\usepackage{vmargin}    % for \setmarginsrb                                      
\usepackage{xspace}
\usepackage{xcolor}
\usepackage{algorithm2e}
\usepackage{latex-pkg/postscript} % HV's elaboration of epsf
\usepackage[nolineno]{latex-pkg/lgrind}     % for typesetting MSL, Gentle, C, C++, ... code
\usepackage[grey]{latex-pkg/quotchap}
\usepackage{latex-pkg/TexShade}
\usepackage[caption=false]{subfig}
\usepackage{url}
\usepackage{multirow}
\usepackage[bf,rm,medium,compact]{titlesec} % nice section titles
\usepackage{pifont}     % to include symbols from different fonts (e.g.  \rtm)
\usepackage{eurosym}    % for the euro symbol, package by Henrik Theiling
\usepackage{booktabs}   % tables as required by journals
\usepackage{fancyhdr}                % fancy page headers/footers
\usepackage{vmargin}    % for \setmarginsrb         
\usepackage{latex-pkg/postscript} % HV's elaboration of epsf
\usepackage[nolineno]{latex-pkg/lgrind}     % for typesetting MSL, Gentle, C, C++, ... code
\usepackage[grey]{latex-pkg/quotchap}
\usepackage{latex-pkg/TexShade}
\usepackage{longtable}
\usepackage{ltcaption}
\usepackage{natbib}
\usepackage{color,soul}
\tcbuselibrary{theorems}

\newif\iftth
\setmarginsrb{36mm}{28mm}{22mm}{35mm}%
	     {30pt}{20pt}{12pt}{40pt}

\newcommand{\vb}[1]{\verb+#1+}

\newcommand{\beq}{\begin{equation}}
\newcommand{\eeq}{\end{equation}}
\newcommand{\beqa}{\begin{eqnarray}}
\newcommand{\eeqa}{\end{eqnarray}}

\makeatletter
\newcommand\arraybslash{\let\\\@arraycr}
\makeatother

\pagecolor{white}

\makeindex

\def\BGfont{\small\tt}
\def\CMfont{\small\tt}
\def\NOfont{\small\tt}
\def\KWfont{\small\tt}
\def\STfont{\small\tt}

\def\VRfont{\small\tt}

\theoremstyle{plain}
\theorembodyfont{\normalfont}
\newtheorem{mytheorem-0}{\textsc{Theorem}}\numberwithin{mytheorem-0}{section}

\newenvironment{new-theorem-0}
{
\begin{mytheorem-0}
\begin{quote}
}
{
\end{quote}
\end{mytheorem-0}
}

\theoremstyle{plain}
\theorembodyfont{\normalfont}
\newtheorem{mycorollary-0}{\textsc{Corollary}}
\numberwithin{mycorollary-0}{section}

\newenvironment{new-corollary-0}
{
\begin{mycorollary-0}
\begin{quote}
}
{
\end{quote}
\end{mycorollary-0}
}

\theoremstyle{plain}
\theorembodyfont{\normalfont}
\newtheorem{myproposition-0}{\textsc{Proposition}}\numberwithin{myproposition-0}
{section}

\newenvironment{new-proposition-0}
{
\begin{myproposition-0}
\begin{quote}
}
{
\end{quote}
\end{myproposition-0}
}

\theoremstyle{plain}
\theorembodyfont{\normalfont}
\newtheorem{mylemma-0}{\textsc{Lemma}}\numberwithin{mylemma-0}{section}

\newenvironment{new-lemma-0}
{
\begin{mylemma-0}
\begin{quote}
}
{
\end{quote}
\end{mylemma-0}
}

\theoremstyle{plain}
\theorembodyfont{\normalfont}
\newtheorem{mydefinition-0}{\textsc{Definition}}
\numberwithin{mydefinition-0}{section}

\newenvironment{new-definition-0}
{
\begin{mydefinition-0}
\begin{quote}
}
{
\end{quote}
\end{mydefinition-0}
}

\newenvironment{new-proof-0}{
\noindent{\textbf{\textit{Proof.}}}
\begin{quote}
}
{
\end{quote}
}

\newcounter{theoremCounter}
\newenvironment{new-theorem-1}
{% This is the begin code
\stepcounter{theoremCounter}
{\noindent}{\textbf{\textsc{Theorem}}} \arabic{theoremCounter}.
}
{}

\newcounter{corollaryCounter}
\newenvironment{new-corollary-1}
{% This is the begin code
\stepcounter{corollaryCounter}
{\noindent}{\textbf{\textsc{Corollary}}} \arabic{corollaryCounter}.
}
{}

\newcounter{lemmaCounter}
\newenvironment{new-lemma-1}
{% This is the begin code
\stepcounter{lemmaCounter}
{\noindent}{\textbf{\textsc{Lemma}}} \arabic{lemmaCounter}.
}
{}

\newcounter{propositionCounter}
\newenvironment{new-proposition-1}
{% This is the begin code
\stepcounter{propositionCounter}
{\noindent}{\textbf{\textsc{Proposition}}} \arabic{propositionCounter}.
}
{}

\newcounter{definitionCounter}
\newenvironment{new-definition-1}
{
\stepcounter{definitionCounter}
{\noindent}{\textbf{\textsc{Definition}}} \arabic{definitionCounter}.
}
{}

\newenvironment{new-proof-1}{
\noindent{\textbf{\textit{Proof.}}}}
{}

\newenvironment{new-item}%
{\begin{list}{$\bullet$}{%
\topsep0pt \partopsep0pt \parsep0pt \itemsep0.1ex plus 0.1ex minus 0.1ex
\setlength{\leftmargin}{0.8\leftmargin}
\setlength{\labelsep}{0.8\labelsep}}}{\end{list}}

\newenvironment{new-enum1}%
{\begin{list}{\arabic{enum1}.}{\usecounter{enum1}%
\topsep0pt \partopsep0pt \parsep0pt \itemsep0.2ex plus 0.1ex minus 0.1ex
\setlength{\leftmargin}{0.8\leftmargin}
\setlength{\labelsep}{0.8\labelsep}}}{\end{list}}

\newenvironment{new-enum2}%
{\begin{list}{\alph{enum2}.)}{\usecounter{enum2}%
\topsep0pt \partopsep0pt \parsep0pt \itemsep0.3ex plus 0.1ex minus 0.1ex
\setlength{\leftmargin}{0.8\leftmargin}
\setlength{\labelsep}{0.8\labelsep}}}{\end{list}}

\newenvironment{new-enum3}%
{\begin{list}{\roman{enum3}.}{\usecounter{enum3}%
\topsep0pt \partopsep0pt \parsep0pt \itemsep0.2ex plus 0.1ex minus 0.1ex
\setlength{\leftmargin}{0.8\leftmargin}
\setlength{\labelsep}{0.8\labelsep}}}{\end{list}}

\newcounter{enum1}
\newcounter{enum2}
\newcounter{enum3}

\newcolumntype{R}{>{\arraybackslash}X}

\newcolumntype{G}{>{\raggedright}X}

\newcolumntype{F}{>{\raggedleft}X}

\makeatother

\renewcommand{\S}{Section~}

\newcommand{\argmax}{\text{argmax}}
\newcommand{\hilby}{\textit{Hilbert}-\textsc{MLE}\xspace}
\newcommand{\vocab}{\mathcal{V}}
\newcommand{\corpus}{\mathcal{D}}
\newcommand{\hcovec}[1]{|#1\rangle}
\newcommand{\hvec}[1]{\langle #1|}
\newcommand{\hdot}[2]{\langle #1 | #2 \rangle}
\newcommand{\norm}[1]{\|#1\|_2}
\newcommand{\eij}{e^{\hdot{i}{j}}}
\newcommand{\R}{\mathbb{R}}
\newcommand{\expectation}[1]{\mathop{\mathbb{E}}_{#1}}
\newcommand{\hder}{\frac{\partial f_{ij}}{\partial\hdot{i}{j}}\xspace}
\newcommand{\pmi}{\mathrm{PMI}\xspace}
\newcommand{\ppmi}{\mathrm{PPMI}\xspace}

\newcommand{\vectors}{\mathbf{T}}
\newcommand{\covectors}{\mathbf{C}}
\newcommand{\courier}[1]{{\fontfamily{qcr}\selectfont #1}}

\newtcbtheorem[number within=section]{example}{Example}%
{colback=black!5, colframe=black!75, fonttitle=\bfseries}{ex}

\pgfplotsset{compat=1.14}
\usetikzlibrary{arrows}
\usetikzlibrary{positioning}

\begin{document}

\pagestyle{empty} %defines the general contents of the headers and footers (e.g. where the page number will be printed)

%%%%%%%%%%%%%%%%%%%
%
% TITLE PAGE
%
%%%%%%%%%%%%%%%%%%%
\begin{onehalfspacing}
\begin{titlepage}
\begin{center}

\vspace*{0.5cm}

% {\sf\bfseries\LARGE  Data Consistency in Scalable\\
% Multi-tier Architectures}

% {\rm\bfseries\LARGE  Data Consistency}
% \vspace{0.15cm}
% {\rm\bfseries\LARGE  in}
% \vspace{0.15cm}
% {\rm\bfseries\LARGE  Multi-tier and Cloud Applications}

{\sf\bfseries\LARGE Word Embedding Algorithms as Generalized Low Rank Models and their Canonical Form}

\vspace{0.15cm}

{\sf\bfseries\LARGE  }

\vspace{0.15cm}

{\sf\bfseries\LARGE  }

\vspace{1.8cm}

{\large Kian Kenyon-Dean }

\vspace{1cm}

Master of Science

\vspace{1.4cm}

School of Computer Science\\
McGill University\\
Montreal, Qu\'{e}bec, Canada\\

\vspace{1.5cm}

% \date{\today}

April 2019

\vspace{1.4cm}

\noindent
A thesis submitted to McGill University in partial\\
fulfillment of the requirements of the degree of\\
Master of Science

\vspace{1.4cm}

{\small \copyright Kian Kenyon-Dean, 2019 }

\end{center}
\end{titlepage}

\cleardoublepage
\end{onehalfspacing}

%%%%%%%%%%%%%%%%%%%
%
% SPACING
%
%%%%%%%%%%%%%%%%%%%
%
% \singlespacing
% \doublespacing
% \onehalfspacing 
%
% \setstretch{1.8}
%

\onehalfspacing 

% \doublespacing

% ROMAN PAGE NUMBERING
\pagenumbering{roman}
\pagestyle{plain}

\begin{Large}
\end{Large}

%%%%%%%%%%%%%%%%%%%
%
% ABSTRACT
%
%%%%%%%%%%%%%%%%%%%
\chapter*{\rm\bfseries Abstract}
Word embedding algorithms produce very reliable feature representations of words that are used by neural network models across a constantly growing multitude of NLP tasks \citep{goldberg2016primer}. As such, it is imperative for NLP practitioners to understand how their word representations are produced, and why they are so impactful. 

The present work presents the \textit{Simple Embedder} framework, generalizing the state-of-the-art existing word embedding algorithms (including \textit{Word2vec} (SGNS) and \textit{GloVe}) under the umbrella of \textit{generalized low rank models} \citep{udell2016generalized}. We derive that \textit{both} of these algorithms attempt to produce embedding inner products that approximate \textit{pointwise mutual information} (PMI) statistics in the corpus. Once cast as Simple Embedders, comparison of these models reveals that these successful embedders all resemble a straightforward \textit{maximum likelihood estimate} (MLE) of the PMI parametrized by the inner product (between embeddings). This MLE induces our proposed novel word embedding model, \hilby, as the canonical representative of the Simple Embedder framework. 

We empirically compare these algorithms with evaluations on 17 different datasets. \hilby consistently observes second-best performance on every extrinsic evaluation (news classification, sentiment analysis, POS-tagging, and supersense tagging), while the first-best model depends varying on the task. Moreover, \hilby consistently observes the least variance in results with respect to the random initialization of the weights in bidirectional LSTMs. Our empirical results demonstrate that \hilby is a very consistent word embedding algorithm that can be reliably integrated into existing NLP systems to obtain high-quality results.

% This work proposes the Simple Embedder framework to generalize several existing word embedding algorithms within the \textit{generalized low rank model} family of matrix factorization algorithms \citep{udell2016generalized}. 

% Our algorithm is theoretically the most parsimonious of the algorithms within the Simple Embedder framework. 
% \hilby substantially reduces the number of hyperparameters that are required for the other algorithms to work.
% Moreover, \hilby is consistently at the high end of the characteristic level of performance of these models across 12 intrinsic and 5 extrinsic evaluation tasks. In particular, \hilby consistently observes second-best performance on every extrinsic evaluation (news-article classification, sentiment analysis, POS tagging on the WSJ and Brown corpora, and supersense tagging), while the first-best model depends varying on the task. Moreover, \hilby consistently observes the least variance in results with respect to the random initialization of the weights in bidirectional LSTMs. These empirical results suggest that \hilby is a highly consistent word embedding algorithm that can be reliably integrated into existing NLP systems to obtain high-quality results.

\clearpage

%\bigskip \bigskip

\chapter*{\rm\bfseries Abr\'eg\'e}

Les algorithmes de plongement lexical (word embedding) produisent des repr\'esentations de mots tr\`es fiables utilis\'ees par les mod\`eles de r\'eseau neuronal dans une multitude de t\^aches de traitement automatique du langage naturel (TALN) \citep {goldberg2016primer}.
En tant que tel, il est imp\'eratif que les praticiens en TALN comprennent comment leurs repr\'esentations de mots sont produites et pourquoi elles ont un tel impact.

Le pr\'esent travail de recherche pr\'esente le cadre Simple Embedder, g\'en\'eralisant les algorithmes de pointe de plongement lexical (y compris Word2vec et GloVe) dans le cadre de mod\`eles g\'en\'eralis\'es de bas rang \citep{udell2016generalized}. 
Nous en d\'eduisons que ces deux algorithmes tentent de produire des produits scalaires des (vecteur d') embeddings qui approchent des statistiques d'information mutuelle ponctuelle (PMI) dans le corpus.
Une fois pos\'es dans le cadre Simple Embedder, la comparaison de ces mod\`eles r\'ev\`ele que ces algorithmes de plongement ressemblent tous \`a une simple estimation du maximum de vraisemblance (MLE) de la PMI param\'etr\'ee par le produit scalaire (des embeddings).
Cette estimation engendre notre nouveau mod\`ele de plongement lexical, Hilbert-MLE, comme \'etant repr\'esentant canonique du cadre Simple Embedder.

Nous comparons empiriquement ces algorithmes avec des \'evaluations sur 17 ensembles de donn\'ees diff\'erents.
Hilbert-MLE observe syst\'ematiquement la deuxi\`eme meilleure performance pour chaque \'evaluation extrins\`eque (classification des nouvelles, analyse des sentiments, \'etiquetage morpho-syntaxique et \'etiquetage supersense), tandis que le meilleur mod\`ele d\'epend de la t\^ache.
De plus, Hilbert-MLE observe syst\'ematiquement la plus faible variance dans les r\'esultats quant \`a l'initialisation al\'eatoire des poids dans les r\`eseaux LSTM bidirectionnels.
Nos r\'esultats empiriques d\'emontrent que Hilbert-MLE est un algorithme de plongement lexical tr\`es coh\'erent pouvant \^etre int\'egr\'e de mani\`ere fiable dans les syst\`emes existants de TALN pour obtenir des r\'esultats de haute qualit\'e.

\cleardoublepage

\chapter*{\rm\bfseries Acknowledgements}

This work would not have been possible without collaboration with Edward Newell -- to whom the originality of the ideas within this work must be attributed -- and to him I am indebted as a scientific colleague and a friend in the writing of this thesis. Additionally, I am eternally indebted to my trusted colleague and advisor, professor Jackie Chi Kit Cheung, who given me consistent guidance throughout my research career as an undergraduate and Master's student. I must also acknowledge professor Doina Precup, who saw promise within me as a researcher when I was just a 2$^{nd}$ year undergraduate taking her introductory programming course; she was my first and continued advisor as a machine learning practitioner. Lastly, I must thank the McGill Computer Science department, the Reasoning and Learning Lab, Mila\footnote{The newly established Qu\'{e}bec Artificial Intelligence Institute.}, NSERC\footnote{The Natural Sciences and Engineering Research Council of Canada.}, the FRQNT\footnote{The Fonds de Recherche Nature et Technologies of Qu\'{e}bec.}, and Compute Canada\footnote{With regional partner Calcul Qu\'{e}bec (\url{https://www.computecanada.ca}).} for giving me consistent funding, resources, and for providing collective expertise of my fellow labmates, without any of which this work would not have been possible.

\cleardoublepage

%%%%%%%%%%%%%%%%%%%
%
%  Table Of Content
%  List Of Figures
%  List Of Tables
%
%%%%%%%%%%%%%%%%%%%

%\setstretch{0.0}
% \begin{singlespacing}
% \begin{spacing}{0.5}
% KZ WORKING BEFORE AUG 11 \setstretch{0.5}

\setstretch{0.7}

\tableofcontents
% \end{spacing}
% \listoffigures
% \end{singlespacing}
% \begin{onehalfspacing}
% \listoftables
% %\listofalgorithms
% \end{onehalfspacing}
% \cleardoublepage

% INCLUDE HVs DEFAULT FORMATTING

\renewcommand{\chaptermark}[1]{\markboth{#1}{#1}}

\renewcommand{\sectionmark}[1]{\markright{\thesection\ #1}}

%\fancyhead[RO,LE]{\sf\bfseries\thepage} %http://www.ctan.org/tex-archive/macros/latex/contrib/fancyhdr/fancyhdr.pdf

%
% \lhead[\fancyplain{}{\sf\bfseries\thepage}]%
%  {\fancyplain{}{\sf\bfseries\rightmark}}
% \rhead[\fancyplain{}{\sf\bfseries\leftmark}]%
%  {\fancyplain{}{\sf\bfseries\thepage}}
%

\lhead[\fancyplain{}{\rm\bfseries}]%
 {\fancyplain{}{\rm\bfseries\rightmark}}
\rhead[\fancyplain{}{\rm\bfseries\leftmark}]%
 {\fancyplain{}{\rm\bfseries}}
% \cfoot{}
\renewenvironment{itemize}%
  {\begin{list}{$\bullet$}%
               {\setlength{\parsep}{0.1cm}
                \setlength{\itemsep}{0.cm}}}%
  {\end{list}}

\renewenvironment{enumerate}%
  {\begin{list}{\addtocounter{enumi}{1}\arabic{enumi}.}%
               {\setlength{\parsep}{0.1cm}
                \setlength{\itemsep}{0.cm}}}%
  {\end{list}\setcounter{enumi}{0}}

\renewcommand{\chapterheadstartvskip}{}

\interfootnotelinepenalty=10000 %priority, to make sure the footnote is not dvided between pages.

\onehalfspacing

%%%%%%%%%%%%%%%%%%%%%%%%%%%%%%%%%%%%%%
%
% BEGIN_Chapters
%
%%%%%%%%%%%%%%%%%%%%%%%%%%%%%%%%%%%%%%

%%%%%%%%%%%%%%%%%%%%
% MTT
%%%%%%%%%%%%%%%%%%%%

\newpage
% ARABIC PAGE NUMBERING
\pagenumbering{arabic}
\pagestyle{fancyplain}

\chapter{\rm\bfseries Introduction}
\label{ch:introduction}

An artificial intelligence practitioner working at the intersection of machine learning and natural language processing (NLP) is confronted with the problem of representing human language in a way that is understandable for a machine. Just as a human cannot naturally comprehend the meaning behind a large vector of floating point numbers, a machine cannot understand the intended meaning behind a set of alpha-numeric symbols that compose the things we call ``words''.

It is therefore left to algorithm designers to determine the most appropriate way to produce meaningful word representations for machines. 
In lexical semantics, one of the most influential guiding principles for designing such algorithms is the \textit{distributional hypothesis}:
\begin{quote}
    You shall know a word by the company it keeps. \citep{firth1957synopsis}
\end{quote}
In other words, words that appear in similar contexts are likely to possess similar meanings. An early influential attempt to algorithmically reflect this principle was the \textit{hyperspace analogue to language} (HAL) \citep{lund1996producing}. The algorithm was straightforward: given a corpus of text, count the number of times words $i$ and $j$ co-occur with each other, $N_{ij}$. After counting these corpus statistics over all $n$ unique words, a word $i$ would be assigned a vector in $\textbf{w}^{(i)} \in \R^n$, where each element holds the cooccurrence count, $w^{(i)}_{j} = N_{ij}$. This method successfully produced word representations with semantic qualities; for example, the authors found that the top 3 nearest neighbors to the vector $\textbf{w}^{(\text{cardboard})}$ were the vectors for the words ``plastic'', ``rubber'', and ``glass'' --- out of 70,000 possible words!

Over the past 23 years since \citeauthor{lund1996producing}'s work, NLP practitioners have been able to implement the distributional hypothesis in ways far beyond a nearest-neighbor analysis in the lexical vector space. Moreover, practitioners have been able to tackle some of the main problems of the distributional hypothesis in recent years \citep{faruqui2015retrofitting,mrkvsic2016counter}. For example, it is often the case that antonyms will appear in very similar contexts (e.g., ``east'' and ``west'', ``cat'' and ``dog''), in which case a direct adherence to the distributional hypothesis may create word representations that do not properly reflect certain semantic relationships between words.

However, the most urgent problem that had to be solved following HAL was the fact that 140,000-dimensional vectors \citep{burgess1998simple} are not very useful in practice, especially not when used as input for downstream NLP tasks (e.g., in sentiment analysis classifiers). Numerous methods were introduced to substantially reduce the dimensionality of word representations: from matrix-factorization based methods using SVD \citep{rohde2006improved}, to learning them within deep neural network language models (NNLMs) \citep{bengio2003neural,collobert2008unified}. Such models were unable to enter into ubiquity due to either their high memory requirments (SVD), or their extremely long training times (NNLMs, on the order of months). It would take \cite{turian2010word} to introduce and motivate a formalization of these types of representations as \textit{word embeddings}. 

Word embeddings are dense, real-valued vector representations of words within a pre-established vocabulary, $\vocab$, induced by some corpus of text, $\corpus$. That is, $d$-dimensional\footnote{Typical models use dimensionalities $d$ between $50$ and $500$.} feature-vector representations of words, $\textbf{v} \in \R^d$, where $d \ll |\vocab|$. These are called \textit{distributed} representations. The elements of the vectors do not correspond to a single semantic concept; one cannot pinpoint the precise interpretation of each element $v_j$ of $\textbf{v}$, unlike with HAL, where each element $w_j$ of $\textbf{w}$ corresponds exactly to the cooccurrence count with respect to word $j$. Instead, the $v_j$ elements of word embeddings represent abstract latent features that reflect some linear combination of the $d$ distributed basis vectors that characterize the embedding space. The corpus that word embeddings are trained on contains a distribution of words that the embeddings either implicitly or explicitly capture, but the question of how to capture this distribution is left to the algorithm designer. If this is done improperly, however, one can produce a set of completely ineffectual embeddings.

Fortunately, large advances were made by \textit{Word2vec} \citep{mikolov2013efficient,mikolov2013distributed} and \textit{GloVe} \citep{pennington2014glove}\footnote{The \textit{Word2vec} papers \citep{mikolov2013efficient,mikolov2013distributed} and \textit{GloVe} \citep{pennington2014glove} collectively have over 25,000 citations (according to Google Scholar) at the time of writing, 6 years after they were released.}. 
These algorithms --- and the predecessors that inspired them \citep{collobert2008unified,turian2010word,mnih2012fast} --- produce word embeddings which constitute the critical building blocks of contemporary deep learning systems for natural language processing (NLP) \citep{collobert2011natural,kim2014convolutional,huang2015bidirectional,goldberg2016primer,peters2018deep}. 
Deep learning systems use these word embeddings as input to obtain state-of-the-art results in many of the most difficult problems in NLP, including (but certainly not limited to): automatic machine translation \citep{qi2018and}, question-answering \citep{rajpurkar2018know}, natural language inference \citep{bowman2015large}, coreference resolution \citep{lee2017end}, and many problems in sentiment analysis \citep{zhang2018deep}. In my work, word embeddings have proven to be indispensable features for various NLP problems. This includes problems like news-article text classification \citep{kenyon2018clustering} and Twitter Sentiment Analysis \citep{kenyon2018sentiment}, to more particular ones such as verb phrase ellipsis resolution \citep{kenyon2016verb} and event coreference resolution \citep{kenyon2018resolving}.

The general applicability and ubiquity of word embeddings warrants a serious theoretical examination. That is, a precise examination of the learning objective that is used to produce the word embeddings. The examination provided in this work yields that both \textit{Word2vec} (SGNS) \citep{mikolov2013distributed} and \textit{GloVe} \citep{pennington2014glove} produce embeddings that, through their dot products, attempt to approximate a slight variant of the \textit{pointwise mutual information} (PMI) between word probability distributions in a corpus. More formally, given two words $i,j$, these algorithms are trained to produce two vectors $\hvec{i}$ and $\hcovec{j}$ such that their inner product approximates their PMI statistic computed from the corpus: $\hdot{i}{j} \approx \pmi(i,j) = \log \frac{P(i,j)}{P(i)P(j)}$. 

In other words, the word embeddings produced by the most widely used embedding algorithms in NLP are trained to predict how ``surprising'' it is to see two words appearing together: how different is the joint probability, $P(i,j)$, from what it would be if the words were independent, $P(i)P(j)$? 
For example, the words ``computational'' and ``linguistics'' probably have a high PMI with each other in an NLP textbook, as they often accompany each other as a noun phrase. On the other hand, words with $\pmi(i,j)=0$ are independent of each other; e.g., ``the'' likely has a PMI of zero with most nouns, it is so common that it will lack a predictive capacity. 

This work is not the first to notice the link between these algorithms and PMI.
\cite{levy2014neural} noticed this as well, providing a critical theoretical leap forward in the understanding of word embeddings. 
They discovered that the language modelling-based sampling method of SGNS (\textit{Word2vec}) is implicitly equivalent to the seemingly distinct method of factorizing a term-context matrix filled with corpus-level cooccurrence statistics. Their main finding was that the SGNS objective is minimized when the inner product between a word vector and a context vector equals the PMI between the word and context in the corpus, minus a global constant. While they found the ultimate objective of SGNS, they could not derive the corresponding loss function for matrix factorization.

% Yet, it is not generally recognized that the two most widely used word embedding algorithms (SGNS) \citep{mikolov2013distributed} and \textit{GloVe} \citep{pennington2014glove} --- \textit{both} produce embeddings that, through their dot products, attempt to approximate a slight variant of the \textit{pointwise mutual information} (PMI) between word probability distributions in a corpus.

\subsubsection{Summary of Contributions in this Work}

The present work offers a theoretical orientation that generalizes SGNS and GloVe into the \textit{Simple Embedder matrix factorization framework}. This framework emerged from intuitions relating embeddings with Hilbert spaces due to the explicit duality present in every word embedding algorithm: each actually produces \textit{two} embeddings for every word, a \textit{term} vector and a \textit{context} vector. The fundamental principle of the Simple Embedder framework is that the model learns term vectors $\hvec{i}$ and context vectors $\hcovec{j}$ such that an objective function is minimized when their dot product equals the value of a predefined association metric $\phi_{ij}$. We prove, for both SGNS and GloVe, that this association metric is a slight variant of the pointwise mutual information (PMI) between words $i$ and $j$ in the corpus on which the algorithm is trained. From this framework, we derive the correct matrix factorization loss functions for both algorithms. 

Using the Simple Embedder framework as the starting point, we additionally derive a novel, canonical word embedding algorithm --- \hilby --- based on the maximum likelihood estimation of the binomial distribution on corpus statistics, finding $\phi_{ij} = \pmi(i,j)$. Deriving the \hilby loss function from the first principles of PMI results in a solution that elegantly handles when $\pmi(i,j) = -\infty$ (which occurs quite frequently) via the exponential function. This elegance in handling PMI is in stark contrast with past approaches to building PMI-based word embeddings, like Singular Value Decomposition. Such methods are forced by circumstance to resort to using positive-PMI\footnote{Positive-PMI (PPMI) is defined as $\ppmi(i,j) = \max(0, \pmi(i,j))$.} \citep{levy2014neural,levy2015improving}, which problematically has little theoretical justification and effectively throws out the majority of the co-occurrence data.
Another advantage of \hilby is that it uses only one empirical hyperparameter (a temperature $\tau$), while SGNS and GloVe both require several empirically tuned hyperparameters and design decisions in order to work properly%
\footnote{SGNS requires the following hyperparameters to be tuned: frequent word undersampling probability $t$, context distribution smoothing value $\alpha$, number of negative samples $k$. GloVe requires two hyperparameters within the empirical weighting function, $\alpha$ and $X_{max}$; it also requires learning biases during training, and term- and context-vector averaging after training as design components. We discuss these throughout the rest of this work; see \cite{levy2015improving} for an in-depth discussion on these and other hyperparameters.}%
. The fact that \hilby does not have to rely on empirical intuitions (unlike SGNS and GloVe) in order to perform well is evidence that our theory is well-grounded.

To empirically verify our theoretical findings, we present experiments using five different sets of word embeddings all trained on a 5.4 billion word corpus. We experiment with: the original SGNS and GloVe algorithms on this corpus, our proposed matrix factorization formulations of these (MF-SGNS and MF-GloVe), and \hilby.
We present results for all five sets of embeddings across 12 intrinsic evaluation datasets, 2 text classification datasets (including news and sentiment classification), and 3 sequence labelling datasets (including part-of-speech tagging and supersense tagging). The latter 5 extrinsic evaluations experiment with prediction models of varying complexity, from logistic regression to deep bi-directional LSTM neural networks. Our experiments also include qualitative analysis and insights into certain properties of the produced \textit{term} and \textit{context} vectors. 

Overall, we find that no one set of embeddings is objectively superior to all others. Analysis of our results and the produced embeddings suggest that MF-SGNS and MF-GloVe correctly and reliably re-implement the original algorithms, often offering better performance. Intrinsic evaluations reveal that \hilby exhibits the standard analogical and similarity-based linear properties of word embeddings, and qualitative analysis offers interesting novel perspectives elucidating what kind of information is expressed in the embeddings' dot products. We also find that \hilby is consistently the second-best model on every extrinsic evaluation task with LSTMs, while the first-best varies depending on the task. \hilby generally observes the least variance in accuracy with respect to the random initialization of LSTM parameters. Our results suggest that \hilby is a very reliable and consistent word embedding algorithm and that it can be incorporated into existing NLP models to improve performance on various tasks.

% \subsubsection{Interpretations of Word Embeddings}

% Each interpretation is equivalent, and 

% Semantic memory vectors that capture lexical semantic components of the language

% Probability models \citep{bengio2003neural,}

% Distributed feature representations \citep{mikolov2013distributed}

% Low-rank estimates of a matrix \citep{levy2014neural,pennington2014glove}

% The cognitive scientist in computational linguistics may describe it as learning a representational model of semantic memory \citep{lund1996producing,burgess1998simple}. The machine learning practitioner might call this distributed representation learning \citep{bengio2003neural,mikolov2013distributed}. In natural language processing, this learning problem is generally described as word embedding \citep{collobert2008unified,collobert2011natural,levy2015improving}.

\section{Statement of Author's Contributions}
This work in this thesis is inspired by and based on a journal submission (currently under review) pursued with equal authorship contribution with Edward Newell, PhD. The \textit{Simple Embedder framework} was conceived by Edward, along with the proofs for SGNS and GloVe as matrix factorization. Additionally, Edward conceived the \hilby algorithm. 
Without Edward, this work would not have been possible. Additionally, Professor Jackie Chi Kit Cheung provided supervision throughout the entirety of this work, and he has been my invaluable Master's supervisor throughout my degree.

My contributions are the following: the engagement and research on the related work (especially with regard to word embedding evaluation); an overview and introduction to the fundamental concepts of this work, particularly with respect to the \textit{pointwise mutual information} metric; discovering certain minorly important formulations of the \hilby loss function theoretically (relating to the marginal probabilities) and empirically (relating to an implementation detail that required algebraic manipulation to work properly); large-scale experimentation and hyperparameter tuning of the learning rates for our MF models; the entirety of the intrinsic and extrinsic experiments across 17 different datasets; and (within a software package that we developed collectively) I initiated and architected the implementation of matrix factorization within automatic differentiation software for our three algorithms (MF-GloVe, MF-SGNS, \hilby).

Note that, in this work, \hilby is derived based on the \textit{binomial} distribution, but it can also be derived from the \textit{multinomial} distribution (as in the journal submission), which is a generalization of the binomial. Such a derivation yields that the approximation in Equation~\ref{eq:characteristicmle} becomes precise; i.e., the denominator $\frac{1}{1-\hat{p}_ij}$ no longer appears when you differentiate the multinomial likelihood function.
\chapter{Word Embeddings in NLP}

% \todo{Define what is a word embedding}
% \todo{Define the transfer-learning motivation}

A word embedding algorithm can be understood as a method to transition from a distribution of discrete sparse vectors to (much) lower-dimensional continuous vectors.
Many perspectives describe how computational models can learn to transition from discrete to continuous representations of language. 
The cognitive scientist in computational linguistics may describe it as learning a representational model of semantic memory \citep{lund1996producing,burgess1998simple}. The machine learning practitioner might call this distributed representation learning \citep{bengio2003neural,mikolov2013distributed}. In natural language processing, this learning problem is generally described as word embedding \citep{collobert2008unified,turian2010word,collobert2011natural,levy2015improving}.

Any word embedding algorithm is dependent on the data it is provided; e.g., embeddings of Twitter data will inevitably possess different qualities than those produced from reading Wikipedia. Despite the algorithm for producing the embeddings --- from sampling-based language models \citep{bengio2003neural,mikolov2013efficient} to matrix factorization \citep{levy2014neural,pennington2014glove} --- the meaning of a word will be defined, according to the \textit{distributional hypothesis}, by the company it keeps \citep{firth1957synopsis}.
As such, word embedding algorithms are not limited to words. The generality of their algorithmic structure allows them to be used to embed objects as diverse as word morphemes \citep{joulin2017bag} and Java code \citep{alon2019code2vec}. 
Even ignoring the generality of word embedding algorithms in other domains, word embeddings are now ubiquitous across NLP \citep{goldberg2016primer}; we thus provide an overview of the main existing algorithms.

% The neural language model approach to producing word embeddings largely neglected the older matrix factorization approaches inspired LSA due to their many shortcomings. This began with the seminal work of \cite{bengio2003neural}. This was further developed upon by \citeauthor{collobert2008unified} in two ground-breaking works (\citeyear{collobert2008unified,collobert2011natural}). \cite{turian2010word} reviewed the existing word embedding methods at the time, finding that those induced from clusters on the Brown corpus could outperform the C\&W embeddings. HPCA was good \cite{lebret2014word}. \cite{mnih2012fast} introduced noise-contrastive estimation to approximate the softmax in the word embeddings, drastically improving learning speed. Thus begins the qualitative leap.

%%%%%%%%%%%%%%%%%%%%%%%%%%%%%%%%%%%%%%%%%%%
\section{Word Embeddings as Language Features}

The primary difficulty in NLP is dealing with the fact that language appears in the form of discrete representations, upon which it is difficult to apply mathematical models. That is, while an image is composed of pixels with meaningful numerical values (the pixel intensity, etc.), a word is simply a string of characters. The first step in any NLP task that requires mathematical modelling, then, is to represent words with numerical feature vectors. 

The most simple way to build feature vectors for words is with a one-hot-encoding. Assuming we have a vocabulary $\vocab$ composed of $|\vocab|$ words, then a feature vector for a word $w$ will simply be a vector $\textbf{v}_w \in \R^{|\vocab|}$ filled with $|\vocab|-1$ zeros, and one $1$ that corresponds to the index of $w$ in the vocabulary array. These representations, however, are of little value on their own since they possess no semantic properties. Consider a vector for the word ``cat'', $\textbf{v}_{\textit{cat}}$, and a vector for the word ``kitten'', $\textbf{v}_{\textit{kitten}}$: these vectors for semantically related words have exactly the same dot product ($\textbf{v}_{\textit{cat}} \cdot \textbf{v}_{\textit{kitten}} = 0$) as the vectors of completely unrelated words (e.g., $\textbf{v}_{\textit{cat}} \cdot \textbf{v}_{\textit{speaker}} = 0$). Not only do these representations lack semantic meaning, but they are also so high dimensional that they are too big to be effectively used as features within deep learning systems. Fortunately, word embeddings, vectors in $\R^d$ ($d \ll |\vocab|$) typically possess much more useful qualities.

The fundamental justification for using word embeddings is that they possess linguistically expressive qualities in relation to each other, unlike one-hot-encoding vectors. Linguistic expressivity can encompass many different notions, depending on the task at hand. For example, it may be desirable to have word embeddings that capture syntax for POS-tagging, while word embeddings that capture sentiment information would be useful for sentiment analysis problems \citep{tang2014learning}. Fortunately, as we discuss throughout the rest of this work, the objective of designing word embeddings that capture the \textit{pointwise mutual information} (PMI) between two words has turned out to be a task that is generically useful for a wide variety of NLP tasks. Indeed, word embeddings are used as the foundational input features of words across many different deep learning architectures \citep{goldberg2016primer}. Below, we briefly show how word embeddings can be used in standard NLP problems; see \citep{goldberg2016primer,young2018recent,galassi2019attention} for thorough overviews of deep learning architectures in NLP (e.g., sequence-to-sequence models).

\subsubsection{Word embeddings for classification}
In a classification problem, we seek to categorize some sample $x \in \mathcal{X}$ as belonging to one of $n$ categories $y_1 \ldots y_n \in \mathcal{Y}$. In NLP, this task is typically characterized by the samples $x$ being arbitrary-length sequences of text; for example, in sentiment analysis $x$ may be a product review and the categories may be $y_1=1$ and $y_2=-1$, corresponding to a ``positive'' versus a ``negative'' review, respectively. 

Any classifier must ultimately be able to predict a category of a sample, $x$, from a fixed length vector representing the sample, $\textbf{v}_x$. This preliminary \textit{representation} mapping can be anything from a simple deterministic feature extraction to a large series of learned non-linear transformations given by a deep neural network. Regardless of how the representation is made, from $\textbf{v}_x$, a classifier must learn some function $f(\textbf{v}_x)$ to perform the final \textit{prediction} mapping of the sample to a category  \citep{kenyon2018clustering}. 

In NLP we are confronted with the difficulty that our samples may have different sequence lengths, inducing the nontrivial problem of determining how to do the \textit{representation} mapping, $x \rightarrow \textbf{v}_x$. Fortunately, there are a large variety of ways to deal with this fact when word embeddings are used as the feature representations of words. Recurrent neural networks (RNNs), popularized in the connectionist literature by \cite{williams1989learning} and \cite{elman1990finding}, have been the ``go-to'' pooling mechanism in neural network solutions for NLP \citep{lawrence1995natural,goldberg2016primer}\footnote{Although today it seems possible that attention mechanisms may surpass RNNs \citep{vaswani2017attention}.}. This is due to the natural inductive bias imposed in their architecture; i.e., reading sentences like humans. \cite{schuster1997bidirectional} introduced bidirectional RNNs (BiRNNs), which consists of two RNNs that read the text from left to right and right to left, which were demonstrated to significantly improve performance of standard RNNs, and are successfully used in many NLP tasks, e.g., sequence labelling \citep{huang2015bidirectional}. Most importantly, \cite{hochreiter1997long} introduced long short-term memory neural networks (LSTMs), which were decisive in establishing RNNs as reliable learning algorithms. LSTMs are likely the most widely used RNN variants as they can successfully mitigate the vanishing and exploding gradient problems faced by RNNs. They have inspired related architectures such as GRUs \citep{chung2014empirical}. 

Let the sequence $x$ be composed of words $w_i$, $x = w_1 \ldots w_L$, where $L$ is the length of the sequence. Letting $\hvec{w_i} \in \R^d$ be the word embedding for word $w_i$, we can do any of the following to produce a fixed-length representation vector, in increasing order of complexity:
\begin{itemize}
    \item \textit{Mean-pooling}: $\textbf{v}_x = \frac{1}{L} \sum_{i=1}^L \hvec{w_i}$, \citep{shen2018baseline}.
    \item \textit{Max-pooling}: $\textbf{v}_x = \text{Max-pool}(\ldots \hvec{w_i} \ldots)$, which stacks the word embeddings into a matrix and selects the max value from each of the $d$ columns \citep{shen2018baseline}.
    \item \textit{CNN-pooling}: $\textbf{v}_x = \text{CNN}(\ldots \hvec{w_i} \ldots)$, which trains a convolutional neural network to dynamically learn how to combine word embeddings for the task \citep{kim2014convolutional}.
    \item \textit{RNN-pooling}: an RNN produces a state vector $\textbf{h}_i$ for every word $w_i$; thus, we could have $\textbf{v}_x = \textbf{h}_L$, or $\textbf{v}_x = \text{Max-pool}(\ldots \textbf{h}_i \ldots)$.
    \item \textit{BiRNN-pooling}: a bidirectional RNN produces two state vectors for every word $w_i$, the forward RNN state $\overrightarrow{\textbf{h}_i}$ and the backward RNN state $\overleftarrow{\textbf{h}_i}$; thus, we can concatenate %
    $\textbf{v}_x = [\overleftarrow{\textbf{h}_1};\overrightarrow{\textbf{h}_L}]$, or %
    $\textbf{v}_x = \text{Max-pool}( 
    \ldots 
        [\overleftarrow{\textbf{h}_i};
        \overrightarrow{\textbf{h}_i}]
    \ldots)$, \citep{conneau2018what}.
\end{itemize}

Any of the above methods, and many more, can be used to produce fixed length vector representations of a sequence of words. 
Regardless of the pooling mechanism used, given any fixed length representation, the classifier will be followed by a logistic regression or feed-forward neural network model, $f(\textbf{v}_x)$, that learns to map $\textbf{v}_x$ to a category label $y \in \mathcal{Y}$. In this work, we experiment with \textit{Max-pooling} as input to logistic regression, \textit{Max-pooling} as input to a feed-forward neural network, and \textit{BiRNN-pooling} using forward-backward concatenation of LSTM hidden states (%
$\textbf{v}_x = [\overleftarrow{\textbf{h}_1};\overrightarrow{\textbf{h}_L}]$) as input to logistic regression
in our classification experiments (\S\ref{sec:empirical}).

\subsubsection{Word embeddings for sequence labelling}
Sequence labelling problems are common in NLP. One standard problem is POS-tagging, the problem of assigning a syntactic part-of-speech label $y \in \mathcal{Y}$ (e.g., noun, verb, etc.) to each word $w_i$ in a sequence $x$. Traditional models, such as conditional random fields (CRF) and hidden Markov models, have typically fared quite well in this task without word embeddings \citep{yao2014recurrent}. However, using word embeddings in collaboration with character-based feature representations in deep BiLSTM models (optionally with a CRF layer) now observes state-of-the-art results for many sequence labelling problems \citep{huang2015bidirectional,bohnet2018morphosyntactic}. In this work (\S\ref{sec:empirical}), we use a vanilla BiLSTM model that takes as input a word embedding, $\hvec{w_i}$, and concatenates the forward and backward hidden states to produce a representation of that word, 
$\textbf{v}_{i} = [\overleftarrow{\textbf{h}_i};\overrightarrow{\textbf{h}_i}]$,
from which label prediction is performed with logistic regression, as in \cite{huang2015bidirectional}.

% \todo{
% Explain how a word embedding is a vector representation of a word from a language.
% Explain how these are used as the foundational features for deep learning systems. That is, how does a deep learning system explicitly use word embeddings as features. E.g., from simple methods like Logistic Regression via embedding pooling, to RNNs, to CNNs, to being inputs for transformers.
% }
% \cite{kim2014convolutional} conclusively showed that word embeddings are a crucial component for deep learning with NLP, finding that results always improve for sentence classification when using word embeddings as features, rather than learning them from scratch for each task.
% \cite{goldberg2016primer} provides a thorough overview of deep learning for NLP, where pre-trained word embeddings are shown to systematically improve results for NLP tasks.

%%%%%%%%%%%%%%%%%%%%%%%%%%%%%%%%%%%%%%%%%%%
\section{Standard Word Embedding Algorithms}
In this section we provide an overview of the seminal algorithms from \cite{mikolov2013efficient,mikolov2013distributed} and \cite{pennington2014glove} --- Word2vec and GloVe. These algorithms lifted word embeddings into ubiquity, and they are now used as feature inputs within a countless number of state-of-the-art NLP systems. Given the ubiquity of their usage, it is desirable to understand precisely how these algorithms produce word embeddings.
% \cite{kim2014convolutional} conclusively showed that word embeddings are a crucial component for deep learning with NLP, finding that results always improve for sentence classification when using word embeddings as features, rather than learning them from scratch for each task.
% \cite{goldberg2016primer} provides a thorough overview of deep learning for NLP, where pre-trained word embeddings are shown to systematically improve results for NLP tasks.

\subsection{Word2vec}

\textit{Word2vec} was introduced by \cite{mikolov2013efficient} in the form of two algorithms, each with certain advantages and disadvantages: the continuous bag-of-words model (CBOW), which predicts the current word given the context, and the continuous skip-gram model, which predicts the surrounding context given the current word. These models were designed with the motivation of substantially improving the computational complexity of the algorithms, along with the general quality of word vectors. 

The standard neural-network-based methods of that time were largely based on inefficient deep feed-forward (or recurrent) neural network language models (NNLMs) \citep{bengio2003neural,collobert2008unified,collobert2011natural}, which would typically take weeks or even months to train. \citeauthor{mikolov2013efficient}'s experiments involved comparing running times between models and comparison of embedding performance across syntactic and semantic analogy completion tasks, finding that skip-gram embeddings performed better on semantic analogies, while CBOW was generally faster and performed better for syntactic analogies.

 \citeauthor{mikolov2013efficient}'s skip-gram model required the use of the hierarchical softmax algorithm to predict the surrounding words, which, despite being logarithmic in the vocabulary size, was still a substantial bottleneck in training time. Soon after introducing skip-gram, \cite{mikolov2013distributed} discovered a much more rapid alternative to hierarchical softmax: negative sampling. This new algorithm, \textit{Skip-gram with Negative Sampling} (SGNS) would become the most widely used word embedding algorithm in NLP over the next few years\footnote{According to Google Scholar at the time of writing, the SGNS paper \citep{mikolov2013distributed} has over 11,500 citations --- more than any other paper on word embeddings --- with \cite{mikolov2013efficient} in second place with over 9,500 citations.}. As one of our main contributions in this work is discovering the correct matrix factorization formulation of SGNS (\S\ref{sec:mfsgns}), we provide an in-depth summary of the algorithm and its motivations below.

\subsubsection{Origins of skip-gram with negative sampling}
The theoretical perspective of skip-gram begins with a standard probabilistic motivation in language modelling, to maximize the average log probabilities of the observed corpus of text data $\corpus$. In effect, this is making an independence assumption about the log-likelihood of the data $\mathcal{L}_\corpus$, but this is naturally quite necessary in NLP in order to avoid the computationally impossible problem of modelling the complete joint distribution of the entire corpus \citep{bengio2003neural}. The likelihood is thus defined as:
\begin{equation} \label{eq:sg-likelihood}
    \mathcal{L}_\corpus = \frac{1}{|\corpus|} \sum_{t=1}^{|\corpus|} \sum_{j \in C(t)} \log P_{\theta}(w_{t+j} | w_{t}),
\end{equation}
where $C(t)$ is a function yielding the context window around the current word $W_t$ (typically, it will yield the 5 to 10 words before and after $w_t$), and $P_{\theta}(w_{t+j} | w_{t})$ represents the probability of the context given the current term as determined by learnable model parameters $\theta$ (i.e., the word embeddings). Note that the difference between skip-gram and CBOW is that CBOW is based on the probability of the term given the context by replacing the probability above with $P_{\theta}(w_{t} | w_{t+j})$.

\paragraph{Softmax.} Modelling this probability requires two sets of vectors to be learned, called by \cite{mikolov2013distributed} as ``input'' and ``output'' vectors. We call these the \textit{context} and \textit{term} vectors, respectively. We use Dirac notation to indicate them, with $\hvec{w_t}$ as the term vector for word $w_t$, and $\hcovec{w_{t+j}}$ as the context vector for word $w_{t+j}$, with their dot product represented as $\hdot{w_t}{w_{t+j}}$. The final output of the embedding model will be the term vectors. The most obvious way to obtain the probability from the model is to use the softmax operation, yielding:
\begin{equation} \label{eq:softmax}
    P_{\theta}(w_{t+j} | w_{t}) = \frac{ \exp{\hdot{w_t}{w_{t+j}}} }{ \sum_{w \in \vocab} \exp{ \hdot{w_t}{w} } },
\end{equation}
where we sum over each word in the vocabulary $\vocab$ in the denominator. Unfortunately, this is an extremely impractical solution due to the need to compute the gradient $\nabla P_{\theta}$ for every element of the summation, requiring $O(|\vocab|)$ evaluations. 

\paragraph{Hierarchical softmax.} Methods such as hierarchical softmax \citep{morin2005hierarchical} have been proposed to substantially improve the computational complexity of the softmax operation to only requiring $O(\log_2 |\vocab|)$ evaluations. However, the structure of the binary tree representation of the output layer can have a strong impact on downstream performance and requires more design decisions and model engineering to work properly. Hierarchical softmax does \textit{not} create context vectors for every word (unlike normal softmax); it instead only uses vectors that represent every inner node in the binary tree \citep{mikolov2013distributed}. Hierarchical softmax is a common choice in recent state-of-the-art representation learning or language modelling systems, such as in \textsc{DeepWalk} for graph vertex embedding \citep{perozzi2014deepwalk}, and it has been improved to \textit{adaptive} hierarchical softmax \citep{grave2017efficient} for state-of-the-art deep language models \citep{dauphin2017language}. 

\paragraph{NCE.} An alternative to hierarchical softmax is Noise Contrastive Estimation (NCE) for language modelling \citep{mnih2012fast,mnih2013learning}, which approximately maximizes the log probability captured by the softmax operation in Equation~\ref{eq:softmax}. The high-level view of NCE is that it seeks to train vectors such that they can be used to discriminate between samples drawn from the data distribution versus samples drawn from a noise distribution. This discrimination is performed by essentially using logistic regression via the logistic sigmoid function:
$
    \sigma(x) = \frac{1}{1 + e^{-x}}
$.
The sigmoid function has several relevant properties that we will make use of later: $\sigma(-x) = 1-\sigma(x)$, and, $\sigma(x) \in [0, 1] \, \forall x \in \R$. By combining logistic regression with sampling from the noise distribution, NCE induces a model with complexity independent of the vocabulary size. It has been successfully used in several deep recurrent models for language modelling and speech recognition tasks \citep{chen2015recurrent,zoph2016simple,rao2016noise}.

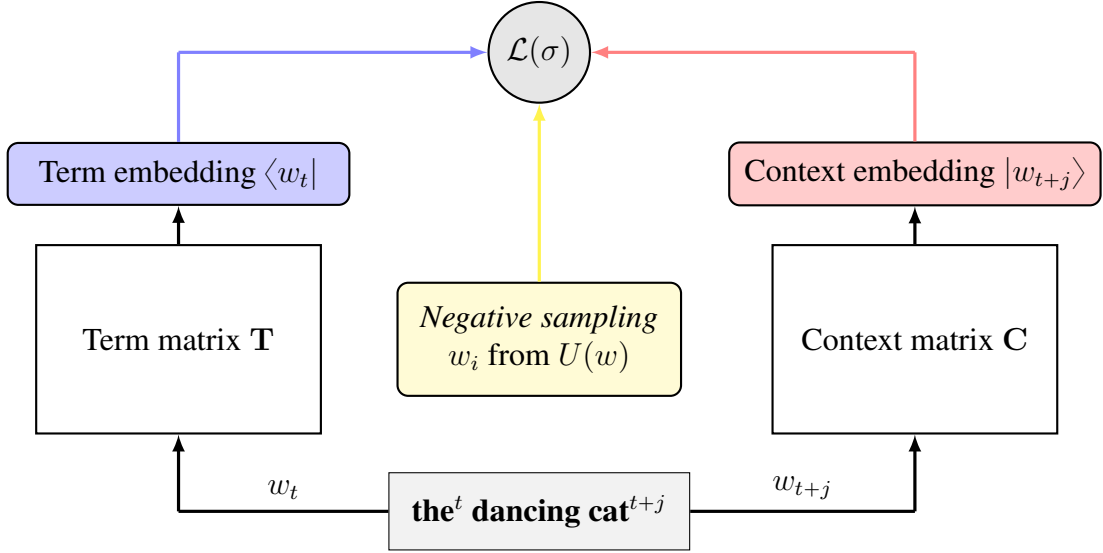
\begin{figure}[t]
\centering
    \begin{tikzpicture}[
  >=latex,
  every matrix/.style={ampersand replacement=\&,column sep=0.5cm,row sep=0.5cm},
  term/.style={draw,thick,rounded corners,fill=blue!20,inner sep=.2cm,minimum width=4.5cm},
  context/.style={draw,thick,rounded corners,fill=red!20,inner sep=.2cm,minimum width=4.5cm},
  embmat/.style={draw,thick,fill=white,inner sep=.3cm,minimum width=3.75cm,minimum height=2.5cm},
  mytextbox/.style={draw,thin,fill=gray!10,inner sep=.3cm},
  gradient/.style={fill=white,inner sep=.3cm},
  empty/.style={inner sep=0cm},
  negsam/.style={draw,thick,rounded corners,fill=yellow!20,inner sep=.3cm,minimum height=1.5cm},
  product/.style={draw,thick,circle,fill=gray!20},
  to/.style={->,very thick},
  to2/.style={-,very thick},
  blueto/.style={->,very thick,color=blue!50},
  blueto2/.style={-,very thick,color=blue!50},
  redto/.style={->,very thick,color=red!50},
  redto2/.style={-,very thick,color=red!50},
  yellowto/.style={->,very thick,color=yellow!80},
  grayto/.style={->,very thick,dotted,color=gray!70},
  grayto2/.style={-,very thick,dotted,color=gray!70},
  every node/.style={align=center}]

  % Position the nodes using a matrix layout
  \matrix{
    % \coordinate (TOPanchorL);\&
    % \node[gradient] (gradient) {Gradient $\nabla$};\&
    % \coordinate (TOPanchorR);\\
%%%
    \coordinate (topanchorL);\& 
    \node[product] (product) {$\mathcal{L} (\sigma)$};\& 
    \coordinate (topanchorR);\\
%%%
    \node[term] (term) {Term embedding $\hvec{w_t}$};
    \&\&
    \node[context] (context) {Context embedding $\hcovec{w_{t+j}}$};\\
%%%
    \node[embmat] (tmat) {Term matrix $\vectors$};\& 
    \node[negsam] (negsam) {\textit{Negative sampling}\\$w_i$ from $U(w)$};\&
    \node[embmat] (cmat) {Context matrix $\covectors$};\\
%%%
    \coordinate (anchorL) {};\&
    \node[mytextbox] (mytextbox) {\textbf{the$^t$ dancing cat$^{t+j}$}};\&
    \coordinate (anchorR) {};\\
  };
  
  % special coordinates to get desired arrow behavior
%   \node[empty,right=of TOPanchorR] (TOPanchorR2) {};
%   \node[empty,below=3.35cm of TOPanchorR2] (CONTEXTANCHOR) {};
%   \node[empty,below=.75cm of CONTEXTANCHOR] (CMATANCHOR1) {};
%   \node[empty,below=.5cm of CMATANCHOR1] (CMATANCHOR2) {};
  
%   \node[empty,left=of TOPanchorL] (TOPanchorL2) {};
%   \node[empty,below=3.35cm of TOPanchorL2] (TERMANCHOR) {};
%   \node[empty,below=.75cm of TERMANCHOR] (TMATANCHOR1) {};
%   \node[empty,below=.5cm of TMATANCHOR1] (TMATANCHOR2) {};
  
  % Draw the arrows between the nodes and label them.
  \draw[to]  (tmat) -- (term);
  \draw[to]  (cmat) -- (context);
  \draw[to2] (mytextbox) -- node[midway,above] {$w_t$} (anchorL);
  \draw[to]  (anchorL) -- (tmat);
  \draw[to2] (mytextbox) --  node[midway,above] {$w_{t+j}$} (anchorR);
  \draw[to]  (anchorR) -- (cmat);
  \draw[blueto2]  (term) -- (topanchorL);
  \draw[blueto]   (topanchorL) -- (product);
  \draw[redto2]   (context) -- (topanchorR);
  \draw[redto]    (topanchorR) -- (product);
  \draw[yellowto] (negsam) -- (product);
%   \draw[grayto2]  (product) -- (gradient);
%   \draw[grayto2]  (gradient) -- (TOPanchorR2);
%   \draw[grayto2]  (gradient) -- (TOPanchorL2);
%   \draw[grayto]   (TOPanchorL2) -- (TERMANCHOR);
  
  % getting arrows between the embedding and matrix
%   \draw[grayto]   (TOPanchorR2) -- (CONTEXTANCHOR);
%   \draw[grayto]   (CMATANCHOR1) -- (CMATANCHOR2);
%   \draw[grayto]   (TOPanchorL2) -- (TERMANCHOR);
%   \draw[grayto]   (TMATANCHOR1) -- (TMATANCHOR2);
  
\end{tikzpicture}
% \end{center}
    \caption{Visual depiction of SGNS on an example sentence ``\textbf{the dancing cat}'', where the current term index $t$ is pointing to ``\textbf{the}'' and the context index $t+j$ is pointing to ``\textbf{cat}''. In the circle, $\mathcal{L} (\sigma)$ abstractly represents Equation~\ref{eq:w2v-orig} as a loss function.}
    \label{fig:my-sgns}
\end{figure}

\paragraph{SGNS.} NCE is bound to a certain probabilistic perspective and requires several complex components in the loss function in order to be theoretically sound. \cite{mikolov2013distributed} was able to substantially simplify it by orienting toward a perspective more based on empirical intuition than probability theory; in their words, ``while NCE can be shown to approximately maximize the log probability of the softmax, the Skip-gram model is only concerned with learning high-quality vector representations, so we are free to simplify NCE as long as the vector representations retain their quality.'' As such, in the data likelihood expression in Equation~\ref{eq:sg-likelihood}, they replace (or, define) the log-posterior probability with the following definition of \textit{skip-gram with negative sampling} (visualized in Figure~\ref{fig:my-sgns}):
\begin{equation} \label{eq:w2v-orig}
    \log P_\theta (w_{t+j} | w_t) := \log \sigma \hdot{w_t}{ w_{t+j}} + \sum_{i=1}^{k} \expectation{w_i \sim U(w)} \log (1 - \sigma \hdot{w_t}{w_i} ),
\end{equation}
where $k$ words $w_i$ are \textit{negative samples} drawn from the unigram noise distribution $U(w)$ (typically, a smoothed distribution, see \S\ref{sec:pmi-corpus}). Despite not actually being a log-probability, it performs quite well as a loss function. The SGNS embeddings are trained on a sample-by-sample basis with stochastic gradient descent. The model receives updates as it scans the corpus $\corpus$ (represented globally in the summation of Equation~\ref{eq:sg-likelihood}). Using dynamic context window weighting (see \S\ref{sec:pmi-corpus} and \cite{levy2015improving}) SGNS extracts the following embeddings at each time step $t$: the term vector $\hvec{w_t}$, the context vector $\hcovec{w_{t+j}}$, and the $i=1\ldots k$ negative samples $\hcovec{w_i}$. This set of $k+2$ vectors will receive updates according to the gradient of Equation~\ref{eq:w2v-orig} with respect to those vectors.

An \textit{attractive-repulsive} interpretation \citep{kenyon2018clustering} can be gained from examining the objective function in Equation~\ref{eq:w2v-orig}. Namely, we observe that the objective will be to \textit{maximize} the term-context inner product $\hdot{w_t}{ w_{t+j} }$ while simultaneously \textit{minimizing} the sum of the term-noise inner products $\hdot{w_t}{w_i}$. Recall that the inner product between two vectors $\textbf{a}$ and $\textbf{b}$ is:  
$
    \textbf{a}^\intercal \textbf{b} = \norm{\textbf{a}} \norm{\textbf{b}} \cos_\theta (\textbf{a}, \textbf{b})
$,
where the $\cos_\theta$ term is bounded between $-1$ and $1$. Thus, the term-context inner product can only be maximized by increasing some combination of the norm of the term vector, the norm of the context vector, and the angle between them (up to co-linearity).
From an optimization perspective, we can see why the term-noise components of the equation are necessary\footnote{I.e., for the same reason as why the denominator in softmax is necessary, see \cite{kenyon2018clustering}.}, as otherwise the model would diverge by maximizing $\hdot{w_t}{ w_{t+j} }$ to infinity (by maximizing the norms of the vectors) as there would be no constraints. 
It is therefore necessary to accumulate a constraining gradient onto the term vector $\hvec{w_t}$, which is effected by the gradient updates according to the partial derivative $\frac{\partial}{\partial \hvec{w_t}} $ of $\log (1 - \sigma \hdot{w_t}{w_i})$. 

The ultimate result is that, at every step, $\hvec{w_t}$ will receive an update that is an accumulation of $k+1$ context vectors: one maximizing update from the true context vector $\hcovec{w_{t+j}}$, and $k$ minimizing updates from the negatively sampled noise context vectors $\hcovec{w_1} \ldots \hcovec{w_k}$. This is why increasing $k$ results in quicker convergence \citep{mikolov2013distributed}, since the term vectors receive more updates at each step. Note also that each of the context vectors will receive one update at a time from $\hvec{w_t}$; the precise behavior of the stochastic gradient descent updates on vectors is described later in \S\ref{sec:simple-mf}.

\subsubsection{Relation to matrix factorization}
\cite{levy2014neural} showed the highly influential result that SGNS is implicitly factorizing a matrix $\textbf{M}$ filled with the pointwise mutual information (PMI) statistics gathered from the corpus. PMI is defined as $\pmi(x,y) = \log P(x,y)/P(x)P(y)$; we will later provide an in-depth engagement and overview of PMI in \S\ref{sec:pmi}.
Formally, \citeauthor{levy2014neural} found that $\textbf{M}_{ij} = \pmi(i,j) - \log k$, where $k$ is the number of negative samples used by SGNS. The consequence of this result is that the SGNS objective is globally minimized when term-context vector dot products equal $\pmi(i,j) - \log k$.  This result has since inspired important related results within the graph embedding literature \citep{qiu2018network}. 

Their proof, however, fell short in two respects. First, they relied on the assumption that each embedding vector was of sufficiently large dimensionality to exactly model the PMI. Second, they did not derive the loss function that would yield this optimization; rather, they relied on using SVD on a different $\textbf{M}$ matrix to try and replicate SGNS, but could not achieve substantive results.
\cite{li2015word}, motivated from a ``representation learning'' perspective, attempted to provide a correct explicit matrix factorization formulation of SGNS, but diverged from \citeauthor{levy2014neural}'s result, arguing that a different matrix was being factorized. In the present work, we find an explicit matrix factorization formulation of SGNS, free of assumptions, that coincides with \citeauthor{levy2014neural}'s result (\S\ref{sec:mfsgns}).

% \subsubsection{SGNS Hyperparameters}
% Frequent word undersampling, context distribution smoothing, negative sampling
% \citep{levy2015improving}

\subsection{GloVe}
\cite{pennington2014glove} proposed Global Vectors (GloVe) as a log bilinear model inspired from two families of algorithms. One the one hand, it incorporates aspects of matrix factorization (MF) approaches, such as the hyperspace analogue to language HAL \citep{lund1996producing,burgess1998simple}. On the other hand, it incorporates the sliding context-window approaches of sampling methods like SGNS \citep{mikolov2013distributed}. In \citeauthor{pennington2014glove}'s presentation of GloVe, it is argued that GloVe takes advantage of the useful parts of MF (i.e., using the global corpus statistics), while simultaneously taking advantage of the sampling methods which implicitly take advantage of data sparsity.

Sampling methods like SGNS are argued to be problematic because of the repetition in the data --- many context windows are going to repeat in the dataset. Therefore, GloVe begins by accumulating the global cooccurrence statistics that would be obtained by a sliding context window applied to the corpus. The term $N_{ij}$ thus refers to the number of times context $j$ occurs with $i$. Notably, while $N_{ij}$ can be understood as a matrix that is of size $|\vocab| \times |\vocab|$, this is a very inefficient representation due to the fact that the majority of elements in the matrix will be $0$, as most term-context pairs simply do not occur. Indeed, \citeauthor{pennington2014glove} find that such zero entries can occupy between 75-95$\%$ of such a data matrix. The authors also acknowledge that infrequent term-context pairs are likely noisy data, and that they probably should not have as much influence on the loss function as more frequent pairs, whose statistics we can be much more confident about. The authors therefore introduce the following weight least-squares loss function in contrast to related methods such as SVD which treats every $(i,j)$ pair equally,
\begin{equation} \label{eq:loss-glv}
    \mathcal{L} = \sum_{(i,j): \, N_{ij}>0} h(N_{ij}) \Big(  \hdot{i}{j} + b_i + \tilde{b}_j - \log N_{ij} \Big)^2,
\end{equation}
where $b_i$ and $\tilde{b}_j$ are learned bias terms for term $i$ and context $j$, respectively. Additionally, they define $h(N_{ij})$ as an empirical weighting function: 
\begin{equation}
h(N_{ij}) = \min\left(1, \left(\frac{N_{ij}}{N_\mathrm{max}}\right)^{\alpha}\right),
\end{equation}
where they tuned these values tuned to $N_{max}=100$ and $\alpha = 0.75$. Note that all cooccurrences more frequent than $N_{max}$ will be weighted equally with this model.

GloVe is efficiently able to mimic matrix factorization without requiring an expensive $N_{ij}$ matrix. It does, however, throw away a substantial amount of data (the zero entries of the matrix, which other methods actually make use of \citep{shazeer2016swivel}). Nonetheless, GloVe has been considerably influential and is used as input in many state-of-the-art NLP systems, including contextualized embedding models such as ELMo \citep{peters2018deep}.

\section{Other Word Embedding Algorithms}
NLP has observed a multitude of different word embedding techniques and algorithms, and it is unfortunate that there not has been a survey of such techniques (although there have been surveys on word embedding evaluation methods \citep{bakarov2018survey}).
We will not provide a complete overview over every existing word embedding algorithm, but we do offer a brief overview of several methods and algorithms that are related to this work.

\paragraph{Matrix representation and factorization.} The first attempts to build word embeddings were motivated from a cognitive science perspective of modelling human semantic memory. \cite{lund1996producing} and \cite{burgess1998simple} propose and examine the Hyperspace Analogue to Language (HAL) method for producing such a model of semantic memory. They used a $n=\,$ 70,000 word vocabulary to construct a $n\times 2n$ matrix of cooccurrence statistics based on using a sliding window across the corpus. In most of their experiments, they do not factorize this matrix, but use very large 140,000 dimensional vectors to represent their words. However, in an analysis provided by \cite{burgess1998simple}, the author used variance-minimization techniques to compress the word vectors to 200 dimensions, which, at that time, was presciently hypothesized to be useful for ``connectionist models''.

Indeed, factorizing this matrix proved to be essential for producing word vectors as features for neural network models. 
Given a term-context matrix filled with corpus statistics $\textbf{M}$, the most obvious way to construct word embeddings is to perform singular value decomposition on the matrix, extracting the vectors corresponding to the top $d$ singular values of the decomposition. Formally, SVD will factorize $\textbf{M}$ into three matrices:
$
    \textbf{M} = U \Sigma V^\intercal,
$
where $U$ and $V$ are orthonormal and $\Sigma$ is a diagonal matrix of the singular values in decreasing order. The outputted term embeddings $\textbf{T}$ will then be the $\alpha$-scaled ($\alpha \in [0, 1]$) vectors corresponding to the top $d$ singular values: $\textbf{T} = U_d \cdot \Sigma_{d}^\alpha$.

This method has been explored by \cite{lebret2014word}, who use the Hellinger distance in collaboration with SVD (PCA) to construct high quality word embeddings. This is in contrast to the implicit distance function being optimized in SVD, which is the Euclidean distance (the mean-squared error). \cite{levy2014neural} and \cite{levy2015improving} have also explored using SVD on cooccurrence statistics to produce word embeddings. Discovering that SGNS \textit{implicitly} factorizes a shifted PMI matrix, they sought to use SVD to \textit{explicitly} factorize such a matrix. Confronted with the problem that $\pmi(0)=-\infty$ (a rather difficult number to factorize with SVD), they resorted to using the \textit{positive} PMI metric: $\ppmi(x) = \max(0, \pmi(x))$. In their experiments, they find that PPMI-SVD generally performs worse than SGNS, although it can sometimes be on par and certain hyperparameter configurations can benefit PPMI-SVD substantially \citep{levy2015improving}.

GloVe is also essentially a matrix factorization technique \citep{pennington2014glove}. Noting that GloVe ignores a huge amount of the data in the cooccurrence matrix (by virtue of such elements being equal to $0$), \cite{shazeer2016swivel} proposed the Swivel algorithm as an alternative to GloVe, which learns word embeddings ``by noticing what's missing''. On the nonzero elements in $\textbf{M}$, they use a standard weighted least-squared error loss with the goal of having term-context vector dot products approximate the PMI. On the zero elements in the matrix, the authors proposed a ``soft hinge'' loss, rather than ignoring them like GloVe. Formally, if $N_{ij}=0$, the loss is:
\begin{equation*}
    \mathcal{L}_{ij} = \log \big[ 1 + \exp(\hdot{i}{j} - \pmi^*(i,j)) \big],
\end{equation*}
where $\pmi^*$ is a $+1$-smoothed variant of PMI that makes it a real number. This algorithm was only shown to be effective in producing embeddings for word similarity and analogy tasks, which is insufficient for as a conclusive evaluation due to the known issues of such evaluations \citep{chiu2016intrinsic,faruqui2016problems,linzen2016issues}. However, they do propose the useful and novel ``sharding'' technique for performing explicit matrix factorization with word embeddings, which we implement in our experiments (Chapter~\ref{ch:empirical}).

\paragraph{Generative model.}
A unifying perspective on PMI-based word embedding algorithms was attempted by \cite{arora2016latent}, who propose a generative model based on ``random walks'' through a latent discourse space. Their arguments and methodology are posed in opposition to \cite{levy2014neural}; \citeauthor{arora2016latent} are particularly opposed to \citeauthor{levy2014neural}'s matrix factorization formulation of SGNS, arguing that their proof that $\hdot{i}{j} \approx \pmi(i,j)$ is insufficient due to their discriminative interpretation of SGNS and model approximation error. They instead propose a generative interpretation, yielding a model with the objective that $\norm{\hvec{i} + \hcovec{j}}^2$ should approximate the log of the cooccurrence statistic between $i$ and $j$. This objective is solved with weighted-SVD. However, they are forced by circumstance of their theoretical perspective to introduce a model that requires more empirically tuned hyperparameters than GloVe or \cite{levy2015improving}'s proposed SVD model (including two for an empirical weighting function and a constant $C$ in the objective). Moreover, their methodology reveals that their algorithm is no better than the existing ones, evaluating only on analogy tasks, which are known to be highly problematic \citep{faruqui2016problems,linzen2016issues}. If their theoretical approach were well-grounded, one would have expected them to derive a theoretically parsimonious model that reduces the number of necessary hyperparameters, and furthermore one would expect that their proposed model would be able to capture analogical structure in the embedding space better than the existing algorithms---neither of these are true.

\paragraph{Subword-based representation.}
FastText \citep{joulin2017bag} learns word embeddings by learning embeddings for character-level N-grams, rather than for the words themselves. This is a particularly useful engineering decision for NLP, as it facilitates elegant and simple handling of out-of-vocabulary (OOV) items in text by constructing an embedding from them by using the character-level N-grams. Moreover, it is likely able to generalize better than standard word embedding, which treats words ``running'' and ``dancing'' as completely different, unrelated elements in a set; FastText, on the other hand, will relate these words through an embedding for the 3-gram ``ing'', which likely would possess features indicating a present tense verb. One could thus interpret FastText as a method that helps to disentangle the factors of variation \citep{bengio2013representation} within a language, allowing certain features of the language to be represented within these morphological N-gram embeddings. However, \cite{joulin2017bag} do not propose a novel loss function nor sampling-method, as it uses SGNS. Therefore, the generalization of SGNS to matrix factorization presented in this thesis extends to FastText. 

Subword information can also be extracted \textit{post-hoc}; that is, from the pre-trained word embeddings themselves, without needing to access the corpus they were trained on. Simple, effective methods have been proposed that effectively capture morphological information within extracted character embeddings. These methods can involve training a character-level LSTM on the pre-trained word embeddings with the objective of ``mimicking'' them \citep{pinter2017mimicking}, and least-squares-based methods have also proven to work quite well in producing such subword embeddings \citep{stratos2017reconstruction,zhao2018generalizing,kim2018learning}. Such methods naturally assist in resolving the OOV problem for pre-trained embeddings. Although, it should be noted that the OOV problem is not so much a problem of not having enough embeddings, but not having the \textit{right} embeddings; indeed, it has been shown across 8 text classification tasks that one can get near-optimal performance with vocabularies smaller than 2500 words (in most cases, less than 1000) \citep{chen2019large}.

\paragraph{Deep contextualized representations.}
In 2018, breakthroughs were made by the models ELMO \citep{peters2018deep} and BERT \citep{devlin2018bert}, which achieved state-of-the-art results on a variety of landmark NLP tasks. Both are deep neural network language models that are trained in a similarly generic way as word embeddings (i.e., with training objectives characterized by predicting context), and are similarly generically useful as input for a variety of NLP tasks.

These models produce contextualized representations of words, ``contextualized'' because the representation of a word is produced as the hidden state of a deep neural network that has seen the surrounding context of the word. This is unlike pre-trained word embeddings, which are un-contextualized at the level of input. For example, a contextualized model would probably produce a more useful representation of the word \textit{bank} in the sentence \textit{she stood near the river bank} than a pre-trained word embedding, as the model would properly handle the polysemous nature of the word, while the pre-trained embedding would likely be overburdened with signal pertaining to the more common occurrence of the word \textit{bank} as a financial institution.

Despite the state-of-the-art results of deep contextualized models, pre-trained word embeddings should not be neglected. Indeed, ELMO in dependent on GloVe embeddings as input during training \citep{peters2018deep}. Moreover, both ELMO and BERT are extremely large deep networks that require anywhere between 8 and 64 GB of GPU RAM --- much larger than the capacity of most desktop computers, let alone mobile devices. Indeed, while these deep models require substantial hardware to even be used in an NLP application, pretrained word embeddings consume dramatically less resources. For example, 40,000 uncompressed pretrained 300-dimensional word embeddings require only around 50MB of memory storage, and can even be efficiently compressed without performance loss to less than 5MB for machine translation and sentiment analysis tasks \citep{shu2017compressing}. 

Lastly, as we shown this work, there are many not well-understood qualities possessed by word embeddings and their algorithms; e.g., the simple canonical embedding algorithm \hilby (\S\ref{sec:hmle}), the relationship between all of the algorithms and PMI-based factorization (\S\ref{sec:comparing}), and the context-recovery powers held by \textit{covectors} (\S\ref{sec:qualitative}). In our view, a strong theoretical understanding of simple word embedding algorithms is a prerequisite for well-informed development of deeper models.

\chapter{Generalizing Word Embedding Algorithms}
In this chapter we will present a generalization of existing word embedding algorithms based on our proposed \textit{simple embedder} framework. In \S\ref{sec:pmi} we provide a review of precisely how corpus statistics are obtained from a corpus of text, the assumptions behind doing so, and what exactly is the \textit{pointwise mutual information} between two words. In \S\ref{sec:simple} we provide the generalization that defines the model class of \textit{simple embedders} as generalized low rank models \citep{udell2016generalized}. In \S\ref{sec:existing} we provide proofs demonstrating why skip-gram with negative sampling \citep{mikolov2013distributed} and GloVe \citep{pennington2014glove} are simple embedders that factorize PMI-based matrices. In \S\ref{sec:hmle} we propose a novel word embedding algorithm, \hilby, based on the principles garnered from analysis of the other algorithms. In \S\ref{sec:comparing} we provide a theoretical comparison between each of the three algorithms, using their loss functions and partial derivatives as the primary points of comparison.

\section{Pointwise Mutual Information (PMI)} \label{sec:pmi}
\cite{levy2014neural} showed the surprising result that SGNS is implicitly factorizing a matrix indexed by a shifted variant of the pointwise mutual information between terms and contexts. We will later show (\S\ref{sec:glove-mf}) that GloVe is also essentially factorizing a PMI matrix. \cite{levy2015improving} show that using SVD to factorize a positive-PMI matrix can obtain comparable results to SGNS and GloVe, under certain conditions. Therefore, the uncanny persistence of PMI across algorithms suggests that it would be worthwhile to revisit it, starting from first principles. The PMI between a \textit{term} $i$ and a \textit{context} $j$ is,
\begin{equation*}
    \pmi(i,j) = \log \frac{P(i,j)}{P(i)P(j)}.
\end{equation*}
Note that $\pmi(i,j) = 0$ if $P(i,j) = P(i)P(j)$; i.e., if $i$ and $j$ are independent of each other. PMI thus offers a measure of association between $i$ and $j$ that computes how ``far'' the \textit{joint probability} is from what it would be under independence. For example, in a standard corpus, we would find that the term ``costa'' has a very high PMI with ``rica'', since the named entity ``costa rica'' is very common, and those words are unlikely to appear on their own. Conversely, it is quite likely that ``politics'' has a negative PMI with ``cereal'', as these two words are completely unrelated and probably appear within different topics. As such, it is not surprising that this measure is the traditional and (often the) best way to measure word association in large corpora \citep{terra2003frequency}. 

The problem, however, is that we do not have direct access to any of the terms in the equation for PMI -- we do not know the true probability distributions for our dataset. Fortunately, with the help of probability theory \citep{bishop2006pattern}, we can obtain highly reliable estimates of the distributions for $P(i)$, $P(j)$, and $P(i,j)$\footnote{In the case of word embedding we have $P(i)=P(j) \,\, \forall i=j$, since we treat terms and contexts equivalently. However, in other uses of simple embedders, there may be different definitions for terms and contexts (e.g., words and POS-tags). Thus, we will treat $P(i)$ and $P(j)$ separately for the purpose of generality.}. We will refer to $P(i)$ and $P(j)$ as the \textit{unigram} (or, \textit{marginal}) probabilities of our terms and contexts (respectively), and we will refer to $P(i,j)$ as the \textit{co-occurrence} (or, \textit{joint}) probability of $i$ and $j$. In the case that the context window is defined as the immediate two words surrounding a term, $P(i,j)$ is the \textit{bigram} probability, but in our case the notion of context can be more broad. 

In \S\ref{sec:statistics} below, we will provide the formal statistical derivation and assumptions behind how we approximate these probabilities, which is not essential for understanding the rest of this work. Informally, we will simply be counting the \textit{number of times} $j$ appears in the context of $i$ for all pairs $(i,j)$, as $N_{ij}$. The joint probability will then be defined as $P(i,j) = \frac{N_{ij}}{N}$, where $N$ is a normalization constant. The unigram probabilities $P(i)$ and $P(j)$ will then simply be obtained by appropriately marginalizing the joint probabilities.

\subsection{Approximating the probability distributions} \label{sec:statistics}
Our goal is to determine a reliable approximation for the joint probabilities of all term-context pairs in order to model PMI. Formally, we presume that $T$ is a discrete random variable representing \textit{terms}, with $|\vocab_T|$ possible values, and $C$ a random variable representing \textit{contexts}, having $|\vocab_C|$ possible values. Our goal is to determine, for each term-context pair $(i,j) \in \vocab_T \times \vocab_C$, the corresponding joint probability $P(T=i, C=j)$. If we can determine the joint probability for every $(i,j)$, then we will have consequently found the marginal probabilities $P(T=i)$ and $P(C=j)$ due to the sum rule for probability distributions on discrete random variables: $P(i)=\sum_j P(i, j)$. 

\begin{figure}[t]
    \centering
    % \begin{center}
\begin{tikzpicture}[
  >=latex,
  every node/.style={align=center},
  every matrix/.style={ampersand replacement=\&,column sep=0.5cm,row sep=0.15cm},
  mytext/.style={inner sep=0.2cm},
  mytextw/.style={inner sep=0.2cm,fill=yellow!25},
  mytextg/.style={inner sep=0.2cm,fill=green!35},
  mybox/.style={inner sep=0.3cm,fill=red!15,very thick},
  ombox/.style={inner sep=0.3cm,fill=blue!15,very thick},
  % edges
  to/.style={->,very thick},
]

%% Matrix for the text.
\matrix{
    \node[mytext] (the) {\courier{the}}; 
    \& \node[mytextw] {\courier{cat}}; 
    \& \node[mytextg] (is) {\courier{is}}; 
    \& \node[mytextw] {\courier{the}};
    \& \node[mytext] {\courier{dancer}};\\
    \&
    \& \node[mytext] (t) {$t$};
    \& \node[mytext] (c) {$c$};
    \& \\
};

%% Context window label
\node[mytext, above=.2cm of is] {\textit{context window}};

%% Node for the (t,c) pair.
\coordinate[right=.55cm of t] (middle);
\node[mybox, below=1.1cm of middle] (pair) {$(t,c)_{5}$};

%% Represent Omega
\coordinate[left=2cm of the] (start);
\node[ombox, below=0cm of start, align=left] (bigrams) {$1.\, (the, cat)$\\$2.\, (cat, the)$\\$3.\, (cat, is)$\\$4. \,(is, cat)$\\$5.\, \ldots$};
\node[mytext, above=0cm of bigrams] {$\mathbf{\Omega}$};
\coordinate[left=4.45cm of pair] (ompointer);

% draw all its edges
\draw[to] (t) -- (pair);
\draw[to] (c) -- (pair);
\draw[to] (pair) -- (ompointer);

\end{tikzpicture}
    \caption{Visualizing how $\Omega$ is constructed when the context window size is $w=1$.}
    \label{fig:window}
\end{figure}
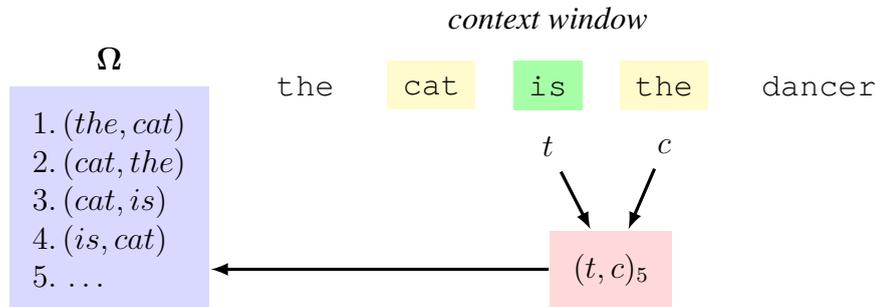

To obtain reliable estimates of these probabilities, we make the strong but traditional (and reliable) assumption in NLP: that our corpus can be interpreted as a (very large) series of identical independent Bernoulli trials \citep{dunning1993accurate}. 
Assume that we are given a corpus $\corpus$ of $|\corpus|$ words, and that the context window for determining term-context co-occurrence is defined as the $w$ words before and after a certain term. 

Now, consider the problem of estimating the joint probability for a \textit{single} term-context pair $(i,j)$. To do this, the precise statistical interpretation is that we first must organize the corpus into a very long list $\Omega$ such that $\Omega$ contains every single term-context pair $(t,c)_x$ observed in $\corpus$ according to $w$, visualized in Figure~\ref{fig:window}. Accordingly, $\Omega$ will have a total of $N=2w|\corpus|$ elements\footnote{Minus a small integer, due to context window clipping at the beginning and end of each corpus document.}.  To estimate $P(i,j)$ from $\Omega$, we can transform $\Omega$ into a series of observed results of $N$ independent Bernoulli trials; that is, a series of $x=1\ldots N$ experiments with \textit{success/failure}\footnote{In the case of word embedding with a symmetric context window, $T$ and $C$ are the same random variable, so $P(i,j)=P(j,i)$; thus, this equality is satisfied if $(t=i \land c=j) \lor (t=j \land c=i)$.} outcomes for the Boolean statement $(t,c)_x = (i,j)$; i.e., does the current term-context pair, $(t,c)_x$, equal the one currently being examined, $(i,j)$?

The \textit{binomial} distribution precisely captures the probability of observing $m$ successes from a series of $N$ Bernoulli trials, where $m \in [0, N]$ is a discrete random variable, and we use $p_{ij}$ as shorthand for $P(i,j)$:
\begin{equation} \label{eq:binomial-dist}
    Bin(m | N, p_{ij}) = \binom{N}{m} p_{ij}^m (1-p_{ij})^{N-m},
\end{equation}
with $\expectation{m} = Np_{ij}$.
In our case, however, we know the number of successes to be exactly $N_{ij}$ since we can simply count them from $\Omega$; but, $p_{ij}$ is unknown. It turns out that $N_{ij}$ is the maximum likelihood estimate of the random variable $m$ \citep{bishop2006pattern}; therefore, because $\expectation{m} = N_{ij} = Np_{ij}$, we have $p_{ij} = \frac{N_{ij}}{N}$ as the maximum likelihood estimate for the co-occurrence probability. 

\subsection{PMI with corpus statistics} \label{sec:pmi-corpus}
Now that we have a reliable estimate of the joint probabilities by using count statistics from the corpus, we can redefine PMI in terms of them. In accordance with the sum rule of probability, we can easily derive the following values for the unigram probabilities as marginals $p_i = P(i)$ and $p_j = P(j)$:
\begin{equation*}
    \mathrm{Let}\,\, N_i = \sum_j N_{ij} \,\,\mathrm{and}\,\, N_j = \sum_i N_{ij}; \,\, \mathrm{then,} \, p_i = \frac{N_i}{N} \,\,\mathrm{and}\,\, p_j = \frac{N_j}{N}.
\end{equation*}
We therefore have everything we need to redefine PMI according to the maximum likelihood estimates of the unigram and joint probabilities garnered from our corpus:
\begin{equation} \label{eq:corpus-pmi}
    \pmi(i,j) = \log \frac{p_{ij}}{p_i p_j} = \log \frac{N_{ij}/N}{(N_i/N)(N_j/N)} = \log \frac{NN_{ij}}{N_i N_j}.
\end{equation}
An important property (or, possibly, drawback) of this formulation is that, if a term-context pair is unobserved, then $N_{ij} = 0$ and thus $\pmi(i,j) = -\infty$. Additionally, note that, if $i$ and $j$ are completely independent, then $p_{ij} = \frac{N_i N_j}{N^2}$. 

\subsubsection{Context window weighting}
The standard word embedding algorithms, \textit{Word2vec} \citep{mikolov2013distributed} and \textit{GloVe} \citep{pennington2014glove} both allow for an arbitrarily sized context window $w$. However, they both also introduce weighting schemes to lessen the impact of seeing a word, say, 10 words away versus one adjacent to the current term. 
This slightly alters how we accumulate $N_{ij}$ from $\Omega$. Namely, rather than properly counting the number of trials, we re-weight the contribution of each trial $x$ according to a weight $r_x \in [0, 1]$ for that trial:
\begin{equation}
    N_{ij} = \sum_{(t,c)_x \in \Omega} r_x \textbf{1}_{(t,c)_x = (i,j)},
\end{equation}
where $\textbf{1}_{\mathrm{bool}}$ is a binary indicator function on the accompanying boolean statement. 
If $r_x = 1 \,\, \forall x$ then this is simply counting up the statistics as before. 
\citeauthor{mikolov2013distributed} instead use \textit{dynamic} context window weighting \citep{levy2015improving}; e.g., if $w=4$ the four contexts following a term would be associated with trials (starting at $x$) with the weights $r_{x}=\frac{4}{4}, r_{x+1}=\frac{3}{4}, r_{x+2}= \frac{2}{4}, r_{x+3}=\frac{1}{4}$. 
Meanwhile, \citeauthor{pennington2014glove} use \textit{harmonic} weighting, where each of those contexts would instead have the weights $\frac{1}{1}, \frac{1}{2}, \frac{1}{3}, \frac{1}{4}$, respectively. \citeauthor{levy2015improving} report experiments when using either technique for weighting, but the difference between the two is inconsequential in terms of downstream performance; our preliminary experimentation described in Chapter~\ref{ch:empirical} also confirms this.

\subsubsection{Context distribution smoothing}
\cite{mikolov2013distributed} use negative sampling to build word embeddings, and draws negative samples according to the unigram context distribution smoothed by an exponent $\alpha=0.75$. \cite{levy2015improving} showed that this has an analog to factorizing the smoothed PMI, $\pmi_{\alpha}$. Below, we present exactly how this smoothing affects the computation of the corpus statistics for the unigram context probability. Note that, by including the exponent, this probability now becomes distinguished from the marginal context probability $p_j$. We therefore must separately raise the $N_c$ frequencies for each context token $c \in \vocab_C$ in order to compute the correct denominator and have a proper probability distribution:
\begin{equation*}
    p^\alpha_j = \frac{N_{j}^{\alpha}}{\sum_{c} N_c^{\alpha}}, \quad \text{thus,} \,\, \pmi_\alpha(i,j) = \log \frac{p_{ij}}{p_{i}p^\alpha_{j}}.
\end{equation*}

% Consider a term $i$ is exactly $x$ tokens from the context $j$; then, \textit{Word2vec} will assign 

%%%%%%%%%%%%%%%%%%%%
\section{Simple Embedders} \label{sec:simple}
To define the class of simple word embedders, we will first introduce notation by informally summarizing their fundamental characteristics before moving on to the exact details in \S\ref{sec:simple-mf}. Our aim is to reason about a set of elements $\vocab$ that appear in some combinatorial structure $\corpus$.  In the case of word embeddings, $\vocab$ is a vocabulary, and $\corpus$ is a text corpus. In other problems where embeddings would be useful, however, $\vocab$ could be entities and relations and $\corpus$ could be a knowledge base, for example.

In some cases it may make sense to define the notion of a \textit{term} and the \textit{context} of a term very differently. For example, in user-item recommendation in may be useful to define a term as a \textit{user} but a context is an \textit{item}. For our case in word embedding, however, we will be using the same vocabulary set $\vocab$ for the terms and contexts. 
Despite having the same vocabulary for terms and contexts, it is still crucial to learn different embeddings for the same words (rather than sharing parameters) analogous to how one would want to learn weight vectors separately from feature vectors in a logistic regression model. Therefore we consider $\vocab$ to comprise two subsets, $\vocab_T$, the \textit{term vocabulary}, and $\vocab_C$, the \textit{context vocabulary}.
Abstractly, a specific embedding problem can be defined as a 4-tuple: 
\begin{equation} \label{eq:4tuple}
   \textbf{Embedding problem:}\,\, (\vocab_T, \vocab_C, \corpus, d),
\end{equation}
where $d$ is the dimension of the target embedding space. 
%The 4-tuple represents a request: ``please embed elements of $\vocab_T$ and $\vocab_C$ as $d$-dimensional vectors, based on how they occur in $\corpus$.''

In turn, an embedder takes an embedding problem as input, and generates two maps, $\vocab_C \longrightarrow \R^d$ and $\vocab_T \longrightarrow \R^d$, by minimizing some global loss function $\mathcal{L}$.
To distinguish the embeddings of $\vocab_T$ from those of $\vocab_C$ we use Dirac notation: $\hvec{i} \in \R^d$ is the term vector associated to term $i \in \vocab_T$; $\hcovec{j} \in \mathbb{R}^d$ is the context vector associated to $j \in \vocab_C$.  In matrix interpretation, $\hvec{i}$ corresponds to a row vector and $\hcovec{j}$ corresponds to a column vector.  Their inner product is represented as $\hdot{i}{j}$.
The logical progression of the simple embedder framework proceeds by taking the global loss function $\mathcal{L}$ and decomposing it into its component-wise losses $f_{ij}$. For word embeddings, each $f_{ij}$ is a loss based on a pairwise relationship between a context $i$ and a term $j$, where there exists an $f_{ij}$ for all term-context pairs $(i,j) \in \vocab_T \times \vocab_C$.

In \S\ref{sec:simple-mf} we will show that a simple embedder problem can be solved with matrix factorization, so long as it exhibits the following three characteristics. The first fundamental characteristic of a simple embedder is that each element-wise loss $f_{ij}$ is only related to other components of the loss function via summation; that is, $\mathcal{L} = \pm \sum_{(i,j)} f_{ij}$. The second characteristic of a simple embedder is that each $f_{ij}$ is a function of exactly two learnable components of the model: the term vector $\hvec{i}$ and context vector $\hcovec{j}$. Indeed, the constraint is even stronger: $f_{ij}$ is a only function of their inner product $\hdot{i}{j}$. The third and final characteristic is that each $f_{ij}$ is minimized (and consequently $\mathcal{L}$ as a whole is minimized) when $\hdot{i}{j} = \phi_{ij}$, where $\phi_{ij}$ is a measure of association between term-context pair $(i,j)$, typically related to their pointwise mutual information. For any simple embedder, the corresponding $\phi_{ij}$ can be found by taking the partial derivative of $\mathcal{L}$ with respect to the dot product $\hdot{i}{j}$ and setting it equal to zero; due to our constraints as described above, this is just the partial derivative on $f_{ij}$; i.e., $\frac{\partial \mathcal{L}}{\partial \hdot{i}{j}} = \frac{\partial f_{ij}}{\partial \hdot{i}{j}}$. 

This simple embedder formalization is inspired by the related work of \cite{udell2016generalized} on \textit{generalized low rank models}. As can be seen in the first equation in Chapter 4 of their work (on generalized loss functions), a simple embedder is a generalized low rank model, where the rank of the implicit matrix is $d$, the dimensionality of the embeddings. Due to this fact, there exists many ways to extend the simple embedder framework that we do not pursue here, including solution methods involving alternating minimization, along with different types of regularization that can be applied upon the embeddings.

\subsection{Simple embedders as low rank matrix factorization} \label{sec:simple-mf}
Embedders differ in terms of the loss function they optimize.  We define the \textit{simple embedders} as those that optimize a loss function that decomposes into a sum of terms wherein model parameters appear only as inner products:
\begin{align} \label{eq:simple}
    \mathcal{L} = \sum_{(i,j) \in \vocab_T \times \vocab_C} 
    f_{ij}\Big(\hdot{i}{j};\phi_{ij}\Big),
\end{align}
where $f_{ij}$ is some function that depends on corpus statistics. Depending on the specific instance of a simple embedder, it may be more appropriate to examine $\mathcal{L} = -\sum_{ij} f_{ij}$ (factoring out a $-1$ from each $f_{ij}$), in which case the loss would be related to the negative log likelihood of the data, rather than a ``least-squares-like'' error function. Either way, $\mathcal{L}$ reaches a global minimum when $\hdot{i}{j}=\phi_{ij}, \, \forall (i,j)$, where $\phi$ is a pre-defined measure of association between $i$ and $j$ based on how they appear in $\corpus$:
\begin{align*}
    \phi: \vocab_T \times \vocab_C \longrightarrow \mathbb{R}.
\end{align*}
Often, $\phi_{ij}$ is a function of the pointwise mutual information (PMI) between words $i$ and $j$ appearing together in a sliding window of width $w$ (a typical $w$ being around 5).

Expanding on the work of \cite{levy2014neural}, we show that any simple embedder -- including SGNS -- can be cast as matrix factorization, and finding an explicit matrix factorization implementation is a matter of formulating the loss function into the form of Eq.~\ref{eq:simple}, and finding the correct $\phi$.
\textit{Low rank} matrix factorization aims to find an approximate decomposition of some matrix $\mathbf{M}$, into the lower rank-$d$ product $\vectors\covectors=\hat{\mathbf{M}}\,{\approx}\,\mathbf{M}$, by minimizing a matrix reconstruction error $\mathcal{L}$ \citep{udell2016generalized}.  
Generalized low rank models perform matrix factorization using a reconstruction error that is computed element-wise on the original and reconstructed matrices: $\mathcal{L} = \sum_{i,j}f_{ij}(\mathbf{M}_{ij}, \hat{\mathbf{M}}_{ij})$, for some element-wise reconstruction error $f_{ij}$ that has a global minimum when $\hat{\mathbf{M}}=\mathbf{M}$.

To see that a simple embedder corresponds to matrix factorization, first let the matrix to be factorized, $\mathbf{M}$, have elements $\mathbf{M}_{ij} = \phi_{ij}$.  Then, pack the term vectors as the rows of matrix $\vectors \in \R^{|\vocab_T| \times d}$, so that $\vectors_{i:} = \hvec{i}$; next, pack the context vectors as the columns of matrix $\covectors \in  \R^{d \times |\vocab_C|}$ so that $\covectors_{:j} = \hcovec{j}$. This means, $\hat{\mathbf{M}}_{ij}=\hdot{i}{j}$, making the loss functions for the simple embedder and low rank matrix factorization identical.

%Therefore, casting SGNS, GloVe, or any embedder as a simple embedder amounts to showing that its loss function takes the form in Eq.~\ref{eq:simple}, and finding the forms of $f$ and $\phi$.

\subsubsection{Gradient descent for simple embedders}
To gain some intuition about the role of $f_{ij}$, it is worth looking at how the factorization problem can be solved with gradient descent.  After initializing term embeddings $\vectors$ and context embeddings $\covectors$ randomly, gradient descent for some learning rate $\eta$ can be performed by looping over the following steps until convergence:
\begin{equation} \label{eq-grad}
\begin{split}
    \hat{\mathbf{M}} &\gets \vectors\covectors \\
    \Delta &\gets {\frac{\partial \mathcal{L}}{\partial\hat{\mathbf{M}}}}\\
    \vectors &\gets \vectors - \eta \Delta \covectors^\intercal\\
    \covectors &\gets \covectors - \eta \vectors^{\intercal}\Delta,
\end{split}
\end{equation}
where the derivative $\frac{\partial\mathcal{L}}{\partial\hat{\mathbf{M}}}$ stands for the matrix of derivatives of $\mathcal{L}$ with respect to each of $\hat{\mathbf{M}}$'s elements:
\begin{equation*}
\Delta_{ij} =\left(\frac{\partial\mathcal{L}}{\partial\hat{\mathbf{M}}}\right)_{ij} = \frac{\partial\mathcal{L}}{\partial\hat{\mathbf{M}}_{ij}}= \hder.
\end{equation*}
The updates to $\vectors$ and $\covectors$ involve adding term vectors to context vectors, and vice versa, in proportion to $\Delta_{ij}=\hder$, because:
\begin{equation*}
\begin{split}
\frac{\partial\mathcal{L}}{\partial\covectors}
= \vectors^{\intercal}\Delta 
\quad \text{and} \quad
\frac{\partial\mathcal{L}}{\partial\vectors} 
= \Delta \covectors^\intercal.
\end{split}
\end{equation*}
Equivalently in vector notation for one $(i,j)$ pair, the update to a given term vector is just $\frac{\partial f}{\partial\hdot{i}{j}}$ times the corresponding context vector, and vice versa:
\begin{equation*}
\begin{split}
\hvec{i} &\gets \hvec{i} - \eta\frac{\partial f}{\partial \hdot{i}{j}}\hcovec{j}^{\intercal}\\
\hcovec{j} &\gets \hcovec{j} - \eta\frac{\partial f}{\partial \hdot{i}{j}}\hvec{i}^{\intercal}.
\end{split}
\end{equation*}
The form of $\hder$ concisely describes the action taken during learning, and we will use this to compare simple embedders in \S\ref{sec:comparing}.  Therefore, we address a specific simple embedder, as a solution to a 4-tuple simple embedder problem (Equation~\ref{eq:4tuple}), by the following 6-tuple that can be solved with any algorithm based on gradient descent: 
\begin{equation} \label{eq:6tuple}
    \textbf{Simple embedder solution:} \,\, (\vocab_T, \vocab_C, \corpus, d; \phi, \hder).
\end{equation}
As described in \S\ref{sec:implementation}, our implementation of the simple embedder solution uses automatic differentiation with \textit{Adam} \citep{kingma2014adam} to perform gradient descent. %However, future work could involve using matrix-factorization solution methods, such as alternating minimization with proximal gradients \citep{udell2016generalized}.

% Four simple embedders described in prior work are exhibited in Table XXX, along with what we propose as the canonical simple embedder, \hilby. For each, we show the form of $\hder$ and the effective value of $\phi$ (the value of $\hdot{i}{j}$ that globally minimizes $\mathcal{L})$.  In the next section, we derive the forms for SGNS and GloVe.

%%%%%%%%%%%%%%%%%%%%%%%%%%%%%%%%%%%%%%%%%%%%%%%%
\section{Existing Algorithms as Simple Embedders} \label{sec:existing}
In this section, we investigate the form SGNS and GloVe when cast as simple embedders.  We find a straightforward interpretation of SGNS as explicit matrix factorization (MF), and discover that GloVe in effect factorizes PMI.

\subsection{Skip-gram with negative sampling (SGNS)} \label{sec:mfsgns}

\cite{levy2014neural} showed that, assuming sufficiently large dimensionality that a given $\hdot{i}{j}$ can assume any value independently of other pairs, SGNS implicitly factorizes the matrix $\mathbf{M}_{ij}=\pmi(i,j) - \log k$, where $k$ is the number of negative samples drawn per positive sample.  However, this assumption is directly violated by the low rank assumption inherent in simple embedders, including SGNS.
\citeauthor{levy2014neural} do not recover $\hder$, and, in its absence, they factorize $\mathbf{M}$ (with negative elements set to 0) using SVD, but this does not yield embeddings equivalent to SGNS. Later, \cite{li2015word} derived an explicit MF form for SGNS, however, their aim was to show that SGNS could be viewed from the perspective of representation learning.  Their derivation results in an algorithm that factorizes cooccurence counts, which obscures the link to \citeauthor{levy2014neural}'s result. \cite{suzuki2015unified} proposed a unified perspective for SGNS and GloVe based on MF, but did not include any experiments to validate their derivation.

Here we present a simple casting of SGNS as matrix factorization, free of additional assumptions, which coincides with the form found by \citeauthor{levy2014neural}.  When compared to other embedders, the form abides by a strikingly consistent pattern, as we will see in \S\ref{sec:comparing}. Moreover, our novel experiments across 17 datasets validate our theoretical findings that our derivation is valid (\S\ref{sec:empirical}).

SGNS \citep{mikolov2013distributed} trains according to the \textit{global}\footnote{Global because SGNS is, in practice, trained stochastically over this full loss using a small mini-batch size when performing updates at each step in the corpus.} loss function\footnote{We use the negative of the objective in \citeauthor{mikolov2013distributed}, taking the convention that $\mathcal{L}$ should be minimized.}:
\begin{equation} \label{eq:sgns-orig}
\mathcal{L} = -\sum_{(t,c)_x \in \Omega}
\left\{
\log \sigma \hdot{t}{c} + \sum_{\ell=1}^k \mathbb{E} \Big[ 
\log (1 - \sigma \hdot{t}{c'_\ell})
\Big]
\right\},
\end{equation}
where $\Omega$ is a list of all $x=1\ldots N$ term-context pairs determined from the context window $w$ in $\corpus$ ($w$ typically around 5), as described in \S\ref{sec:pmi}.  
The term
$\sum_{\ell=1}^k \mathbb{E} \big\{ 
\log (1 - \sigma  \hdot{t}{c'_\ell})
\big\}$ 
is approximated evaluating the summand for $k$ negative samples $c'_1 \ldots c'_k$ drawn from (a smoothed version of) the unigram distribution.  
If we collect all terms with like pairs $(t,c)_x = (i,j)$ together, we obtain:
\begin{equation} \label{eq:sgns-mf-loss}
\mathcal{L} = - \sum_{
    (i,j) \in \vocab_C \times \vocab_T
}
\Big\{
N_{ij}\log \sigma \hdot{i}{j} +  
N_{ij}^- \log( 1- \sigma \hdot{i}{j})
\Big\},
\end{equation}
where $N_{ij}$ is the number of times context $j$ occurs within a window around term $i$ in the corpus (as defined by the context window parameter $w$), and $N_{ij}^-$ is the number of times that context $j$ is drawn as a negative sample for term $i$. Note that this is beginning to take the form of a simple embedder already due to the formulation of the loss function:
\begin{equation*}
    \mathcal{L} = \sum_{(i,j)} f_{ij},\quad \text{where} \,\, f_{ij} = - N_{ij}\log \sigma \hdot{i}{j} - N_{ij}^- \log( 1- \sigma \hdot{i}{j}).
\end{equation*}
However, the simple embedder solution (Equation~\ref{eq:6tuple}) requires the derivative $\hder$ and the optimal value $\phi_{ij}$ in order to fully express SGNS as a simple embedder.
Differentiating $\mathcal{L}$ with respect to $\hdot{i}{j}$ yields:
\begin{equation*}
\frac{\partial\mathcal{L}}{\partial\hdot{i}{j}} = \hder =  
N_{ij}^- \sigma \hdot{i}{j}
- N_{ij} (1 - \sigma \hdot{i}{j}).
\end{equation*}
Proceeding with some algebraic regrouping of terms, we have:
\begin{equation*}
\begin{split}
    \hder &= (N_{ij}^{-} + N_{ij}) \sigma \hdot{i}{j} - N_{ij} \\
    &= (N_{ij}^{-} + N_{ij}) \sigma \hdot{i}{j} - (N_{ij}^{-} + N_{ij}) \sigma \bigg( \log \frac{N_{ij}}{N_{ij}^{-}} \bigg)\\
    &= (N_{ij}^{-} + N_{ij}) \bigg[ \sigma \hdot{i}{j} - \sigma \big( \log \frac{N_{ij}}{N_{ij}^{-}} \big) \bigg],
\end{split}
\end{equation*}
which emerges due to the useful property yielded by the logistic sigmoid function $\sigma (x) = \frac{1}{1+e^{-x}}$, that $a = (a+b)\sigma (\log \frac{a}{b})$.

Now, we can clearly see that the partial derivative $\hder$ will only be equal to zero when $\hdot{i}{j} = \log \frac{N_{ij}}{N_{ij}^{-}}$. Thus, we have discovered a minimum to the loss function $f_{ij}$ for SGNS, allowing us to define the following for our simple embedder:
\begin{equation*}
    \phi_{ij} = \log\frac{N_{ij}}{N_{ij}^-}.
\end{equation*}
Further inspection of this quantity requires us to recall the definition of $N_{ij}^-$. When negative samples are drawn according to the unigram distributions in the corpus, and there are $k$ negative samples drawn for each positive sample, we will have drawn the $(i,j)$ pair as a negative sample $N_{ij}^- = k\frac{N_i N_j}{N}$ times, in expectation. Therefore, by Equation~\ref{eq:corpus-pmi}:
\begin{equation*}
\phi_{ij} = \log\frac{N_{ij}}{N_{ij}^-} = \log\frac{NN_{ij}}{kN_iN_j} = \pmi(i,j) - \log k,
\end{equation*}
which is consistent with the findings of \cite{levy2014neural}. Note, when the unigram distribution for negative sampling is smoothed according to a value $\alpha \in (0, 1)$ (which is often necessary for good results), $\pmi$ simply changes to its smoothed variant, $\pmi_{\alpha}$, which raises each context corpus statistic to the power $\alpha$, as discussed in \S\ref{sec:pmi-corpus}.

Therefore, we propose to train SGNS embeddings by using the loss function defined by Equation~\ref{eq:sgns-mf-loss}. We call this the explicit matrix factorization variant of SGNS, or MF-SGNS. This is unlike \citeauthor{levy2014neural}, who use singular value decomposition to factorize the shifted PMI matrix, which problematically cannot handle when $\pmi(i,j)=-\infty$ by virtue of using a least-squares-like loss function that directly factorizes the matrix. 

Meanwhile, the loss function we have derived handles the $-\infty$ problem as a result of the sigmoid function. Observe that when $\pmi(i,j)=-\infty$ it is because $N_{ij}=0$. So, observing Equation~\ref{eq:sgns-mf-loss}, we see that the loss for such a sample will simply be the easily-differentiable value $f_{ij} = -N_{ij}^{-} \log(1-\sigma\hdot{i}{j})$. More intuitively, if we examine the gradient when $N_{ij}=0$, we see that it will yield the following, since $\sigma(-\infty)=0$:
\begin{equation*}
\begin{split}
    \frac{\partial f_{ij}}{\partial \hdot{i}{j}} &= N_{ij}^{-} \big(\sigma \hdot{i}{j} - \sigma(\log 0)\big)\\
    &= N_{ij}^{-} \sigma \hdot{i}{j}.
\end{split}
\end{equation*}
This is a naturally attenuating gradient whose loss $f_{ij}$ will be minimized as $\hdot{i}{j} \rightarrow -\infty$, or as $\sigma \hdot{i}{j} \rightarrow 0$. Fortunately, as the dot product gets more negative, the gradient itself shrinks toward $0$, leading to an asymptotically decreasing slope already near zero by the time $\hdot{i}{j} = -3$ (since $\sigma(-3)\approx0.05$). In other words, as $\hdot{i}{j} \rightarrow -\infty$ we also observe that $\frac{\partial f_{ij}}{\partial \hdot{i}{j}} \rightarrow 0$, causing the negative growth of the dot product to asymptotically slow down as it gets more negative. In practice, we found that the model learns in a very stable manner (constantly decreasing loss) without diverging.

%%% Glove %%%
\subsection{GloVe} \label{sec:glove-mf}

GloVe's loss function \citep{pennington2014glove} already takes a form closely resembling that of a simple embedder:
\begin{equation*}
\begin{split}
    \mathcal{L} &= \sum_{(i,j) \in \vocab_T \times \vocab_C} f_{ij} \\
    f_{ij} &= 
    \begin{cases}
      h(N_{ij}) \Big(\hdot{i}{j} + b_i + \tilde{b}_j - \log N_{ij} \Big)^2 & N_{ij} > 0 \\
      0 & N_{ij} = 0
    \end{cases}\\
    h(N_{ij}) &= \min\left(1, \left(\frac{N_{ij}}{N_\mathrm{max}}\right)^{\alpha}\right),
\end{split}
\end{equation*}
where $b_i$ and $\tilde{b}_j$ are bias parameters for term $i$ and context $j$, and where $N_{\mathrm{max}}$ and $\alpha$ are hyperparameters.  We write $f_{ij}$ such that it depends only on an inner product of model parameters (i.e., as a simple embedder) by defining:
\begin{equation*}
\begin{split}
\tilde{\hvec{i}} &= \Big[\hvec{i}_1\;\;\,\hvec{i}_2\;\;\,\cdots\;\;\hvec{i}_{d}\;\;b_i\,\,1\,\,\Big] \\
\tilde{\hcovec{j}} &= \Big[\hcovec{j}_1\;\;\hcovec{j}_2\;\;\cdots\;\;\hcovec{j}_{d}\;\;1\;\;\tilde{b}_j\Big]^\intercal\\
\implies f_{ij}&=h(N_{ij})(\tilde{\hdot{i}{j}} - \log N_{ij})^2.
\end{split}
\end{equation*}
From this, we derive the following necessary components for GloVe's simple embedder solution (Equation~\ref{eq:6tuple}):
\begin{equation*}
    \hder = 2 h(N_{ij}) ( \hdot{i}{j} + b_i + \tilde{b}_j - \log N_{ij}), \,\, \text{and}\,\, \phi_{ij} = \log N_{ij} - b_i - \tilde{b}_j.
\end{equation*}

\begin{figure}[t]
    \centering
    \includegraphics[width=0.8\columnwidth]{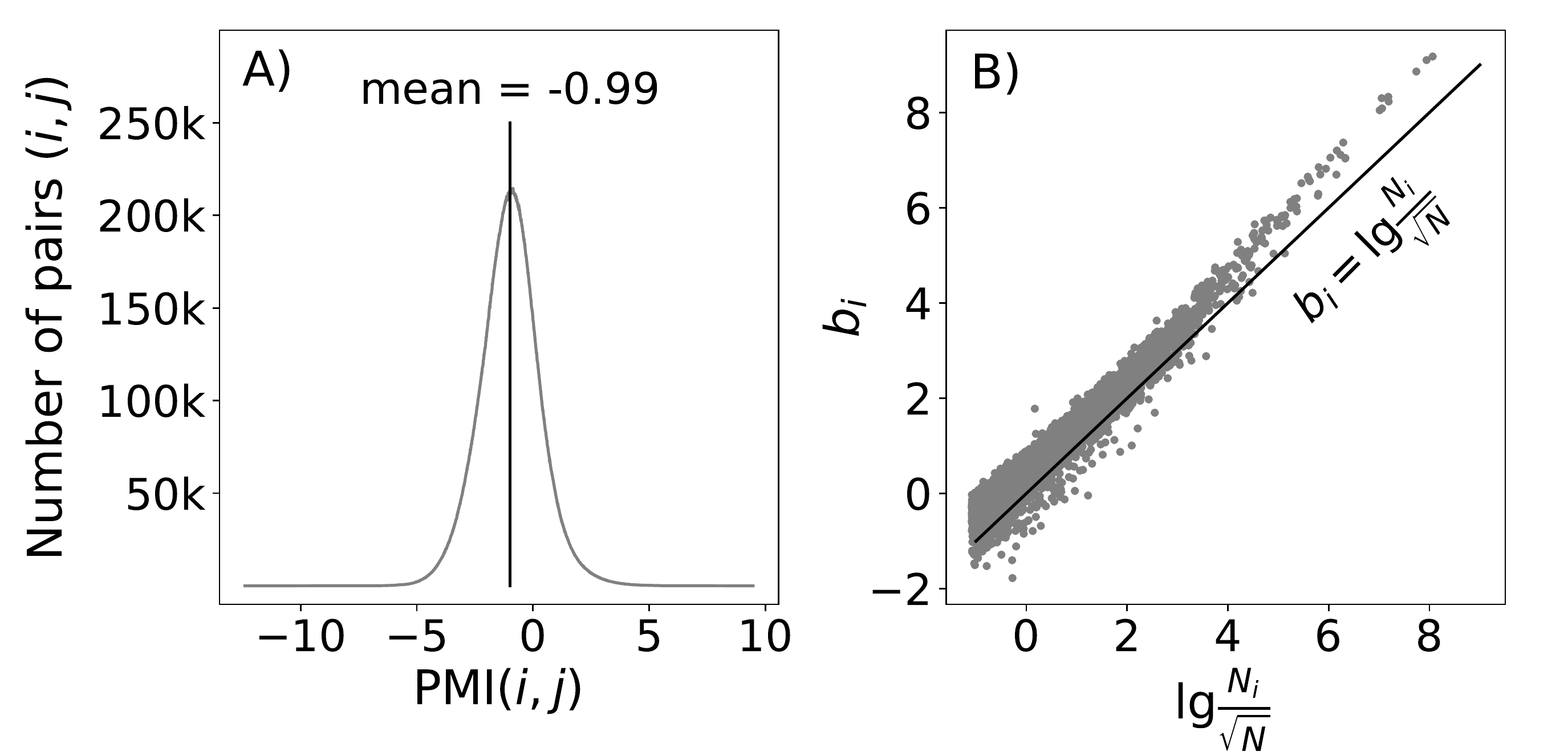}
    \caption{\textbf{A}) Histogram of empirical $\pmi(i,j)$ values, for all pairs $(i,j)\in \vocab_C \times \vocab_T$ for which $N_{ij}>0$ in our corpus.  Note that glove ignores pairs having $N_{ij}=0$.  \textbf{B}) Scatter plot of GloVe's learned biases after $10$ training epochs, using $X_\mathrm{max}=100$ and $\alpha=3/4$ (see \cite{pennington2014glove} for a justification of $X_\mathrm{max}$ and $\alpha$).}
    \label{fig:glove-is-pmi}
\end{figure}

While this demonstrates that GloVe is a simple embedder, it is worthwhile to inspect the role played by the bias terms.
Authors have suggested that the presence of the bias terms might in effect cause $\hdot{i}{j}$ to tend towards PMI during training. \citet{shi2014linking} showed that the correlation of the bias term $b_i$ with $\log N_i$ increases during training up to about 0.8.  \citet{arora2016latent} argued that the bias terms may be equal to $\log N_i$ based on the overall similarity of GloVe to another model in which bias terms with those values appear by design.  We find that the bias terms in fact closely approximate $\log\frac{N_i}{\sqrt{N}}$.  So $\hdot{i}{j}$ minimizes $f_{ij}$ when $\hdot{i}{j} = \log N_{ij} - b_i - \tilde{b}_j \approx \pmi(i,j)$. To see this, set $\hder = 0$:
\begin{align*}
    \hdot{i}{j} + b_i + \tilde{b}_j = \log N_{ij}.
\end{align*}
Multiplying by $1$ and applying the log-rule, we get:
\begin{equation} \label{eq-glove-is-pmi}
\begin{split}
    \hdot{i}{j} + b_i + \tilde{b}_j &= \log \left(\frac{N_iN_j}{N}\frac{N}{N_iN_j}N_{ij}\right) \\
    &= \log \frac{N_i}{\sqrt{N}} + \log\frac{N_j}{\sqrt{N}} + \mathrm{PMI}(i,j) .
\end{split}
\end{equation}
On the right side, we have two terms that depend respectively only on $i$ and $j$, which are candidates for the bias terms.  Based on Eq.~\ref{eq-glove-is-pmi} alone, we cannot draw any conclusions.  We would need to show that these terms accounted for all of the bias, that is, that PMI is centered.  
In fact, PMI \textit{is} nearly centered, see Fig.~\ref{fig:glove-is-pmi}A.  
PMI has an almost normal distribution centered close to 0 (slightly negative).  So, the $\log N_i/\sqrt{N}$ terms absorb nearly all of the bias.
Empirically, we find that the bias terms become close to $\log N_i/\sqrt{N}$ after training, see Fig.~\ref{fig:glove-is-pmi}B. This means that GloVe can be added to the growing list of simple embedders whose objective is closely related to PMI.

% Our goal is to have our model learn the natural binomial distribution of the data. In order to do so, we can make use of those statistics that we can store quite easily in memory ($N$, $N_i$, $N_j$) rather than expensively computing the model's marginal approximations of these statistics at every iteration. We thus treat $N$, $N_i$, $N_j$ as fixed numbers in order to approximate the full likelihood of the $N_{ij}$ co-occurrence statistics.

% It is thus quite advantageous to take advantage of the numbers for which we have reliable estimates ($N_i$, $N_j$, $N$), rather than forcing the model to learn these in addition to the joint probabilities. Moreover, having the model use inner products to exactly model the PMI (and thus indirectly capture the joint probabilties) is very much desirable as it is often in the reasonable range of $-5$ to $5$ (Figure~\ref{fig:glove-is-pmi}A); otherwise, if the inner products approximated $p_{ij}$ directly, the components of the vectors would be extremely small floating point numbers, likely lacking any expressivity at all.

%%%%%%%%%%%%%%%%%%%%%%%%%%%%%%%%%%%%%%%%%%%%%%%%%%%%%%
%%%%%%%%%%%%%%%%%%%%%%%%%%%%%%%%%%%%%%%%%%%%%%%%%%%%%%
\section{\hilby: A Canonical Simple Embedder} \label{sec:hmle}
We now derive \hilby, a canonical simple embedder, from first principles. To begin the derivation, we acknowledge the unanimous choice among simple embedders, intentional and not, to structure the model as estimating PMIs by inner products.  
Next, we acknowledge that any statistics that we calculate from $\corpus$ come with statistical uncertainty.  As is common in statistical machine learning \citep{bishop2006pattern}, we will derive our gradient by maximizing the likelihood of the corpus statistics, subject to this uncertainty. Being consistent with the existing simple embedders, we will set $\phi_{ij} = \pmi(i,j)$ as the target objective for our inner products $\hdot{i}{j}$. Our goal now is to determine what the most appropriate loss function is in order to guide the model into the direction of capturing PMI as well as possible.

\subsection{Likelihood of the data}
To model the PMI statistics, we would like to \textit{maximize the likelihood} of the $N_{ij}$ corpus statistics that define the PMI (Equation~\ref{eq:corpus-pmi}). Our model only needs to model the joint probabilities $p_{ij}$ (as a function of $N_{ij}$) to be effective; it is not necessary to separately model the unigram probabilities ($p_i$ and $p_j$) since they can simply by obtained by marginalizing over the joint probabilities. 
To maximize the likelihood, we need to know the probability distributions that generated our corpus statistics, $N_{ij}$. According to standard probability theory \citep{bishop2006pattern}, a distribution of count statistics can be reliably modelled with a \textit{binomial distribution}, $\mathrm{Bin}(N_{ij} | p_{ij})$ (Equation~\ref{eq:binomial-dist}). Because our $N_{ij}$ statistics are obtained from counting, it is reasonable to assume that every $N_{ij}$ is the parameter of a binomial distribution. From this assumption, we will be able to derive the log-likelihood of the data, $\log \mathcal{L}_{\corpus}$, from the assumed likelihood function, $\mathcal{L}_{\corpus}$:
\begin{equation} \label{eq:loglike}
\begin{split}
    \mathcal{L}_{\corpus} :=& \prod_{i,j} Bin(N_{ij} | p_{ij})\\
    =& \prod_{i,j} \binom{N}{N_{ij}} p_{ij}^{N_{ij}} (1-p_{ij})^{N-N_{ij}}, \quad \therefore\\
    \log \mathcal{L}_{\corpus} =& \sum_{i,j} \bigg[ N_{ij} \log p_{ij} + (N-N_{ij}) \log(1 - p_{ij}) + \log \binom{N}{N_{ij}} \bigg].
\end{split}
\end{equation}

\subsection{Maximizing the likelihood}
For our model to capture the log-likelihood of the data in Equation~\ref{eq:loglike}, we would need for our model's inner products to approximate the joint probabilities, $p_{ij}$. However, we are confronted with the problem that our original motivation was to have the inner products \textit{approximate the PMI}, $\hdot{i}{j} \approx \phi_{ij} = PMI(i,j)$. Fortunately, there is a direct relationship between the PMI and the joint probability that allows us to take advantage of the unigram statistics that can be easily stored in memory:
\begin{equation} \label{eq:phat}
\begin{split}
    \hdot{i}{j} \approx PMI(i,j) &= \log \frac{p_i p_j}{p_{ij}} = \log \frac{N^2}{N_i N_j} p_{ij}, \quad \therefore\\
    \hat{p}_{ij} &= \frac{N_i N_j}{N^2} \eij,
\end{split}
\end{equation}
where we define \textit{the model's approximation of the joint probability} as $\hat{p}_{ij}$. Note how the model takes advantage of the existing unigram statistics in order to form this approximation. It is also important to note that $\hat{p}_{ij}$ is an extremely small floating point number due to the fact that, because a typical corpus is in the order of billions of words, $N^2 \in O(10^{18})$. An alternative to this model design would be to use learned biases to model the unigram probabilities, like GloVe does. However, we have already showed that the biases in GloVe essentially learn the unigram statistics (\S\ref{sec:glove-mf}), so we argue that it is not necessary to learn them separately when we already know what values they should take. Therefore, we must store each value of $N_i$ and $N_j$, along with $N$, in order to train the model. This is computationally inconsequential, as this will only require $2|\vocab| + 1 \in O(|\vocab|)$ additional values to be stored in memory (meanwhile, storing the embeddings uses $O(d|\vocab|)$ memory).

In Equation~\ref{eq:loglike}, we provided the definition of the log-likelihood of the data, $\log \mathcal{L}_{\corpus}$, as a function of the data statistics, $p_{ij}$ and $N_{ij}$. We can now use this definition to inform our derivation of the loss function, $\mathcal{L}$, for \hilby. We will define the loss function assuming that \hilby should model the negative log-likelihood by substituting the actual data, $p_{ij}$, with our model's approximation of the data, $\hat{p}_{ij}$. By minimizing this loss function, we will therefore be maximizing the likelihood of the data as captured by our model.
Note that we remove $\log \binom{N}{N_{ij}}$ since it is a constant not dependent on the model parameters, and we normalize the loss to be independent of the corpus size $N$, which usefully allows us to empirically tune to a corpus-size-independent learning rate:
\begin{equation} \label{eq:hmle-loss}
\begin{split} 
    \mathcal{L} &= 
    - \frac{1}{N} \sum_{i,j} \Big[ N_{ij}\log\hat{p}_{ij} + (N-N_{ij}) \log(1-\hat{p}_{ij}) \Big] \\
    &= - \sum_{i,j} \Big[ p_{ij}\log\hat{p}_{ij} + (1-p_{ij}) \log(1-\hat{p}_{ij}) \Big] \\
    &= - \sum_{i,j} f_{ij}(\hdot{i}{j}; \phi_{ij}), \text{where}, \\
    f_{ij} &:= p_{ij}\log \hat{p}_{ij} + (1-p_{ij})\log(1-\hat{p}_{ij}).
\end{split}
\end{equation}

$\mathcal{L}$ has the form needed for a simple embedder: it depends only on inner products of the model parameters, and, as will be apparent once we take the derivative, it has its optimum at $\hdot{i}{j} = \phi_{ij} = \pmi(i,j)$. Beginning from $f_{ij}$ as defined above, omitting obvious steps, substituting the definition of $\hat{p}_{ij}$ (Equation~\ref{eq:phat}), we obtain the following characteristic gradient for \hilby:
\begin{equation} \label{eq:characteristicmle}
\begin{split}
    \hder &= p_{ij} + (1 - p_{ij}) (-\hat{p}_{ij}) \frac{1}{1 - \hat{p}_{ij}} \\
    &= \bigg( p_{ij} (1-\hat{p}_{ij}) + \hat{p}_{ij}(p_{ij} - 1) \bigg) \frac{1}{1-\hat{p}_{ij}} \\
    &= \big( p_{ij} - \hat{p}_{ij} \big) \frac{1}{1-\hat{p}_{ij}} \\
    &\approx p_{ij} - \hat{p}_{ij}\\
    &= \frac{N_i N_j}{N^2} \big[ e^{\pmi(i,j)} - e^{\hdot{i}{j}} ].
\end{split}
\end{equation}
Note, the approximation obtained when removing the fraction on the right-hand side is reasonable as the denominator is always almost equal to $1$ --- in fact, even the most likely cooccurrence (``of the'') will only ever have a denominator equal to around $0.999$, and every other cooccurrence will denominators much much closer to $1$. 

We have therefore yielded the characteristic\footnote{We call this the \textit{characteristic} gradient as, in practice, we can use automatic differentiation \citep{paszke2017automatic} on the original loss function (Equation~\ref{eq:hmle-loss}). Nonetheless, preliminary experiments revealed no difference in performance between using the ``true'' gradient or this ``characteristic'' gradient, providing evidence that the approximation is appropriate.} gradient of \hilby. Examining this gradient allows for useful insights to be drawn. Namely, we can clearly observe that the gradient is equal to zero when $\hdot{i}{j} = \pmi(i,j)$ and therefore that the objective of this loss function is indeed $\phi_{ij} = \pmi(i,j)$.
This provides a simple and intuitive gradient that causes the estimated cooccurrence probabilities to be drawn toward the empirical cooccurrence probabilities.  When the model parameters are transformed into this domain, no difficulties arise when $N_{ij}=0$ and thus when $\pmi(i,j) = -\infty$, since $e^{-\infty} = 0$, and, as $\hdot{i}{j}$ becomes more negative, the gradient flattens, naturally preventing divergence.

%%%%%%%%%%%%%%%%%%%%%%%%%%%%%%%%%%%%%%%%%%%%%%%%%%%%%%%
\subsection{Accounting for model lack-of-fit}
At this point, by maximizing the likelihood of the data, we are implicitly attributing all discrepancy between model and data to statistical uncertainty of the corpus statistics.  However, the fundamental assumptions of simple embedders, linearity and low rank of $\mathbf{M}$, are thus assumptions about the corpus statistics. These assumptions are useful, but we do not expect actual statistics to be expressible as a low rank system of linear equations.
This leads to a certain amount of lack-of-fit. At some point, it is inevitable that resolving error $\epsilon = \hdot{i}{j} - \phi_{ij}$ for one $(i,j)$ pair will conflict with resolving error in others. This is not statistical error, so it does not make sense to penalize high-frequency pairs more for error arising from lack-of-fit. A priori, error from model lack-of-fit should perhaps be borne more evenly throughout the model.

Ideally, we would compromise between the stratified assignment of error based on likelihood, and the evenly distributed assignment we would otherwise assume in the absence of the statistical treatment.  We draw this compromise by attenuating the multiplier, by raising it to an inverse power, a \textit{temperature} $\tau\geq 1$, 
\begin{equation*}
    \hder = \left( \frac{N_i N_j}{N^2} \right)^{\frac{1}{\tau}} \big[
    e^{\hdot{i}{j}} - e^{\pmi(i,j)}
    \big],
\end{equation*}
which we implement in automatic differentiation by multiplying the original $\mathcal{L}$ by $(\frac{N_i N_j}{N^2})^{\frac{1}{\tau} - 1}$.  Greater $\tau$ distributes responsibility for lack-of-fit more evenly throughout the model, while lesser $\tau$ attributes responsibility to the statistically least grounded (rarest) pairs.
We are unable to derive $\tau$ theoretically, so we introduce it as the only model-specific hyperparameter for \hilby.  
In our tuning experiments (\S\ref{sec:tuning}), we find $\tau=2.0$ to be optimal; i.e., the square root.

%%%%%%%%%%%%%%%%%%%%%%%%%%%%%%%%%%%%%%%%%%%%%%%%%%%%
%%%%%%%%%%%%%%%%%%%%%%%%%%%%%%%%%%%%%%%%%%%%%%%%%%%%
\section{Comparing Simple Embedders} \label{sec:comparing}
In the section, we compare the three simple embedders discussed in this work: SGNS (\S\ref{sec:mfsgns}), GloVe (\S\ref{sec:glove-mf}), and \hilby (\S\ref{sec:hmle}). Recall that the quantity $\frac{N_i N_j}{N^2} = p_i p_j$, and that $p_i p_j$ defines the probability of $(i,j)$ cooccurrence \textit{assuming independence} (\S\ref{sec:pmi}). We therefore define $p^I_{ij} := \frac{N_i N_j}{N^2}$ as this probability of $(i,j)$ cooccurrence under independence. 

We will begin by examining the multiple forms of the single-cell $f_{ij}$ loss functions for each algorithm (see their corresponding sections to refer to the notation); note that we distribute $\frac{1}{N}$ into SGNS to make $f_{ij}$ independent of the corpus size, and for GloVe we define the equivalent multiplier function $h^{*}(p_{ij}) = \min (1, (\frac{N p_{ij}}{N_{max}})^\alpha)$ for the purpose of commonality of exposition:
\begin{equation*}
\begin{split}
% SGNS
\mathrm{SGNS:} \quad  f_{ij} &= p_{ij} \log \sigma \hdot{i}{j} +  k p^{I}_{ij} \log (1 - \sigma \hdot{i}{j})\\
% MLE
\mathrm{MLE}: \quad  f_{ij} &= p_{ij} \log \hat{p}_{ij} +  (1-p_{ij}) \log (1 - \hat{p}_{ij}) \\
% GLV
\mathrm{GloVe:} \quad  f_{ij} &= h^*(p_{ij}) \big( \hdot{i}{j} +  b_i + \tilde{b}_j - \log N_{ij} \big)^2 \\
&\approx  h^*(p_{ij}) \big( \tilde{\hdot{i}{j}} - \pmi(i, j) \big)^2. 
\end{split}
\end{equation*}
We first observe that SGNS and \hilby possess loss functions that strongly resemble the traditional binary cross-entropy loss function in logistic regression. In particular, recall that the sigmoid function allows $\sigma \hdot{i}{j}$ to be interpreted as a probability as it is constrained between $0$ and $1$, which is a common interpretation with logistic regression. Meanwhile, GloVe uses a standard weighted squared error loss function. Despite these very different loss functions, recall that each is minimized when $\hdot{i}{j}$ is equal to some small variant of the pointwise mutual information:
\begin{equation*}
\begin{split}
\mathrm{SGNS:} \quad \phi_{ij} &= \pmi(i,j) - \log k\\
\mathrm{MLE:} \quad \phi_{ij} &= \pmi(i,j) \\
\mathrm{GloVe:} \quad \phi_{ij} &= \log N_{ij} - b_i - \tilde{b}_j \approx \pmi(i,j).\\
\end{split}
\end{equation*}

While each algorithm has approximately the same optimum (based on PMI), the low-rank assumption of simple embedders is likely not valid for real data, and it is thus impossible to have a fully complete approximation of the PMI statistics unless the embedding dimensionality $d = |\vocab|$. Therefore, to understand \textit{how} they are attempting to solve this impossible problem, it is more informative to examine the derivatives of these loss functions:
\begin{equation*}
\begin{split}
% SGNS
\mathrm{SGNS:} \quad  \hder &= (p_{ij} + k p^{I}_{ij}) \bigg[ \sigma \hdot{i}{j} - \sigma \big( \pmi(i,j) - \log k \big) \bigg] \\
% MLE 
\mathrm{MLE}: \quad  \hder &= (p^{I}_{ij})^{\frac{1}{\tau}} \quad\quad\,\, \bigg[ e^{\hdot{i}{j}} - e^{\pmi(i,j)} \bigg] \\
% GLV
\mathrm{GloVe:} \quad  \hder &= 2 h^*(p_{ij}) \quad\,\,\, \bigg[ \hdot{i}{j} + b_i + \tilde{b}_j - \log N_{ij} \bigg] \\
&\approx 2 h^*(p_{ij}) \quad\,\,\, \bigg[ \tilde{\hdot{i}{j}} - \pmi(i, j) \bigg].
\end{split}
\end{equation*}
From this perspective, we can observe that the gradients all bear a very similar update formulation. Namely each gradient takes the form $(\mathrm{multiplier}){\;\cdot\;}(\mathrm{difference})$. The multiplier establishes the relative weight that each $(i,j)$ pair bears on the global loss function.
The difference component defines a notion of displacement between $\hdot{i}{j}$ and its target.

Examining the multiplier components, we consistently see a function that increases sublinearly with either the $(i,j)$ probability under independence (\hilby), with the actual joint probability (GloVe), or by combining both of these (SGNS). The inclusion of such a multiplier was, in some cases \citep{pennington2014glove}, motivated by the idea that, when a cooccurrence count $N_{ij}$ is very small, the relative uncertainty is very large. It is noteworthy that the multiplier of SGNS is also sublinear in $p_{ij}$, (though it is not immediately obvious) because common words are undersampled \citep{mikolov2013distributed}.

Examining the difference components, observe that the loss gradient must be able to handle when $p_{ij} = N_{ij}=0$ (and hence $\pmi$ becomes negative infinity). GloVe does this by simply by having $h^*(0) = 0$, effectively ignoring all of those entries of the matrix --- this decision was largely motivated from an optimization perspective due to the very large size of $|\vocab|^2$ \citep{pennington2014glove}. However, \hilby and SGNS are able to handle the negative infinity in a more manner that is more mathematically elegant than simply zeroing out responsibility. Namely, $\sigma(-\infty) = 0$ for SGNS, and $e^{-\infty} = 0$ for \hilby. Consequently, as $\hdot{i}{j}$ becomes more and more negative for these $p_{ij}=0$ samples, the gradient will naturally attenuate toward $0$ in a stable manner. 

% For example, observe the following SGD updates for the two algorithms for a single $(i,j)$ pair on the term vector $\hvec{i}$ with context vector $\hcovec{j}$:
% \begin{equation*}
% \begin{split}
%     \hvec{i} \leftarrow& \hvec{i} - \hcovec{j}^\intercal  \eta(p_{ij} + k p^I_{ij}) \sigma \hdot{i}{j} \\
%     \hvec{i} \leftarrow& \hvec{i} -  \hcovec{j}^\intercal  \eta(p^I_{ij})^\frac{1}{\tau} e^{\hdot{i}{j}}
% \end{split}
% \end{equation*}
% Interestingly, however, in SGNS the dot products will move in the direction of getting more largely positive (due to $\sigma(\hdot{i}{j}) + 1$ as the difference update), while in MLE the dot products will get in the direction of getting more largely negative (due to $e^{\hdot{i}{j}} - 0$ as the difference update); in both cases nonetheless, the magnitude of the vectors will be increasing.

Overall, three key features emerge as a consistent pattern across the gradients of these well-performing (see \S\ref{sec:empirical}) embedding algorithms:
\begin{itemize}%[itemsep=0mm]
    \item an optimum at some variant of PMI;
    \item a multiplier that is sublinear in $p_{ij}$; and
    \item a difference function that attenuates as $\pmi \to -\infty$.
\end{itemize}
In many cases, ad hoc design elements and tuned hyperparameters are behind these features. For example, the minus $\log k$ shift is required in SGNS, and GloVe requires bias vectors to approximate the PMI. Additionally, SGNS requires negative samples and (implicitly) the undersampling threshold in its multiplier, GloVe requires the empirically tuned weighting function $h$, and, even \hilby needs the $\tau$ parameter to assist in dealing with the very unequal distribution of multipliers across all $(i,j)$ pairs.

In systematic testing of embedders including SGNS and GloVe, \cite{levy2015improving} found that the success of the two models largely depends on careful tuning of hyperparameters, which we discussed in \S\ref{sec:mfsgns} and \S\ref{sec:glove-mf}. 
%Recall that SGNS requires undersampling common words based on a soft threshold parameter $t$, smoothing the unigram distribution used for drawing negative samples using a parameter $\alpha$, and applying what amounts to a negative global shift on PMI values by adjusting the ratio $k$ between the number of negative and positive samples.
The observation that differences in performance mainly originate in hyperparameter settings, together with apparent convergence in the form of the gradients wherein these hyperparameters appear, suggests that these models are collectively tending toward some canonical solution.

We argue that \hilby represents such a canonical solution for several reasons. \textit{First}, it is derived from the basic principles underlying the data distribution of count statistics; i.e., the loss function is derived directly from the binomial distribution on PMI statistics. Conversely, SGNS and GloVe are derived from a perspective more based on intuition than from the actual data distribution. \textit{Second}, \hilby is theoretically and empirically parsimonious; that is, it requires only one hyperparameter and no ad hoc design decisions to function properly, while the others require several such parameters and decisions each. \textit{Third}, as we show in the next chapter, \hilby achieves the characteristic level of success of these models across 17 different evaluation tasks.

\chapter{Empirical Evaluation of Embeddings} \label{ch:empirical}

We evaluate our pretrained embeddings on a wide variety of intrinsic and extrinsic evaluation tasks. The purpose of these experiments is to answer the following questions:
\begin{itemize}
    \item Do our matrix factorization re-implementations of the original GloVe and SGNS algorithms effectively replicate them? In what cases would matrix factorization be preferred over the originals?
    \item Is \hilby, as a canonical embedding algorithm over the others, comparable to the others, in terms of performance? In what cases would \hilby be preferred over the originals?
\end{itemize}
We first provide a high-level overview of the corpus used to produce the embeddings and how the matrix factorization formulations are implemented. Next, we discuss the intrinsic evaluations on word similarity tasks. With the learning rates and embeddings fixed based on performance on the 10 word similarity datasets, we moved on to final evaluation. We present results for analogical reasoning, news article classification, sentiment analysis, POS-tagging, and supersense tagging. We additionally perform a qualitative analysis of the embeddings that elucidates the differences between the term and context vectors. 

Overall, we find that the embedding algorithms tend to achieve similar performance across tasks. While \hilby consistently obtains second best results regardless of the extrinsic evaluation task, the other algorithms trade first place depending on the specific task. We also find that \hilby observes the least sensitivity to the random initialization of the weights in the deep bidirectional long short-term memory neural network (BiLSTM) models \citep{schuster1997bidirectional,huang2015bidirectional}. Our results provide evidence that our matrix factorization implementations of SGNS and GloVe correctly implement the originals. Our results additionally offer empirical evidence to justify our theoretical claims that \hilby is a canonical embedding algorithm because it obtains characteristic performance (often, best or second-best) within the class of simple embedders (i.e., compared with [MF-]GloVe and [MF-]SGNS) across every evaluation task.
% \todo{Factor in Jackie's concern here.}

\section{Practical Implementation} \label{sec:implementation}
% Despite the theoretical guarantees offered by alternating minimization methods \citep{udell2016generalized}, our successful implementation of matrix factorization updates both matrices simultaneously by using existing automatic differentiation software \citep{paszke2017automatic} in conjunction with \textit{Adam} \citep{kingma2014adam} as the stochastic optimization algorithm to perform the gradient descent as described in Equation~\ref{eq-grad} (\S\ref{sec:simple}). 

Our experiments are performed on a corpus ($\corpus$) by merging Gigaword 3 \citep{graff2007english} with a Wikipedia 2018 dump, yielding 5.4 billion tokens. We limit $\vocab_C$ and $\vocab_T$ as the 50,000 most frequent tokens in $\corpus$ to keep the compute time and storage requirements for our large number of experiments reasonable.
We use a $5$-token context window ($w=5$), and have models produce $300$-dimensional embeddings. We use the released implementations and hyperparameter choices of SGNS (number of negative samples $k = 15$, undersampling probability of $t = 10^{-5}$, context distribution smoothing of $0.75$) and GloVe (term and context vector averaging, empirical weighting function with $\alpha = \frac{3}{4}$ and $X_{max} = 100$). 

Our matrix-factorization (MF) reimplementations of the existing algorithms use the same effective hyperparameters\footnote{Note, while GloVe uses \textit{harmonic} context window weighting (\S\ref{sec:pmi-corpus}), we found no difference in performance when using \textit{dynamic} weighting for the matrix factorization reimplementation (MF-GloVe) -- this is consistent with the results of \cite{levy2015improving}; so, for each of our implemented MF models we use dynamic weighting.}; see \cite{levy2015improving} for a discussion of each in detail.
Our MF models (MF-SGNS, MF-GloVe, and \hilby) use PyTorch\footnote{Version 0.4.1, \url{pytorch.org}} to take advantage of GPU-acceleration, automatic differentiation \citep{paszke2017automatic}, and \textit{Adam} \citep{kingma2014adam} as the stochastic optimization algorithm to perform the gradient descent as described in Equation~\ref{eq-grad}. Each was allotted 24 hours for training.

In theory, MF-SGNS and \hilby require dense matrix factorization of the full PMI matrix; in practice, we can implement the matrix by taking advantage of sparsity due to the fact that over 90\% of the values will be equal to $-\infty$. However, the theory still requires\footnote{However, future work would involve determining how to sub-sample these $(i,j)$ pairs in an efficient way, as they are likely not particularly influential on the gradient.} 
gradient descent for the inner products $\hdot{i}{j}$ on such $(i,j)$ pairs, so $\hat{\textbf{M}}$ will still need to be densely computed. As such, to correspond to the theory as precisely as possible, we densely compute this matrix of dot products. In practice, we found that a 12GB GPU can handle when $\hat{\textbf{M}}$ is no larger than 12,500 by 12,500 dimensions. We therefore use sharding over the $\textbf{M}$ matrix, as used in the Swivel matrix factorization algorithm \citep{shazeer2016swivel}. Essentially, this means that a single full batch update across the entire $\textbf{M}$ matrix (i.e., the full dataset) requires sixteen 12,500 by 12,500 shards to be iterated over by scanning through our 50,000 by 50,000 dimensional matrix. Ultimately, this resulted in about 4 seconds per iteration of full batch matrix factorization.

%%%%%%%%%%%%%%%%%%%%%%%%%%%%%%%%%%%%%%%%%%%%%%%%%%
%%%%%%%%%%%%%%%%%%%%%%%%%%%%%%%%%%%%%%%%%%%%%%%%%%
\section{Intrinsic Evaluation} 
\begin{table}[t]
    \centering
    \begin{tabular}{lrll}
    \toprule
        \textbf{Dataset} & \textbf{Coverage} & \text{Acronym} & \text{Related Work}\\
    \midrule
    \textit{Word Similarity}\\
       Baker-Verb 143 & 100$\%$ & B143 & \citep{baker2014unsupervised} \\
       MEN development set & 97.6$\%$ & MENd & \citep{bruni2012distributional} \\
       MEN test set & 96.5$\%$ & MENt & \citep{bruni2012distributional} \\
       Radinsky Mechanical Turk & 99.3$\%$ & RMT & \citep{radinsky2011word} \\
       Rare words dataset & 30.5$\%$ & Rare & \citep{luong2013better} \\ 
       SemEval 2017 & 88.6$\%$ & SE17 & \citep{camacho2017semeval} \\
       Simlex999 & 99.6$\%$ & S999 & \citep{hill2015simlex} \\
       Wordsim353 (relatedness) & 99.6$\%$ & W353R & \citep{agirre2009study} \\
       Wordsim353 (similarity) & 98.0$\%$ & W353S & \citep{finkelstein2002placing}  \\
       Yang-Powers Verb & 88.5$\%$ & Y130 & \citep{yang2006verb} \\ \midrule
    \textit{Analogical Reasoning} \\
       Balanced Analogy Test Set & 68.6$\%$ & BATS & \citep{gladkova2016analogy} \\
       Google Analogy dataset & 87.1$\%$ & Google & \citep{mikolov2013efficient} \\
    \bottomrule
    \end{tabular}
    \caption{Datasets used for word similarity and analogy experiments and the percent coverage obtained with our 50,000 word vocabulary.}
    \label{tab:similarity-datasets}
\end{table} 

Word embeddings evaluation methods are fall into two classes: intrinsic and extrinsic. Intrinsic evaluations seek to evaluate specific qualities in the vectors without requiring any learning nor augmentation of the vectors. These include word similarity ranking and analogical reasoning, as we describe later. Such tasks do not evaluate how well the embeddings can be expected to perform in a downstream machine learning system, and in fact have no correlation with downstream performance \citep{faruqui2016problems,chiu2016intrinsic}. However, it is possible that one might seek to use embeddings for tasks such as analogical recovery, or in a way akin to how embeddings are utilized in collaborative filtering \citep{musto2016learning}. It is therefore still useful to examine performance on intrinsic evaluation tasks. In Table~\ref{tab:similarity-datasets} we enumerate each dataset used for intrinsic evaluation, along with the vocabulary coverage\footnote{By \textit{coverage} we mean the percent of samples in the dataset (2-tuples for word similarity, 4-tuples for analogies) for which each word in the sample was present in our vocabulary.} for each task.

Intrinsic evaluation methods typically rely on the cosine similarity function between two \textit{term} vectors $\hvec{i}$ and $\hvec{j}$:
\begin{equation} \label{eq:cosinesim}
    \cos(\hvec{i},\hvec{j}) = \frac{\hvec{i} \cdot \hvec{j}}{\norm{ \hvec{i}} \cdot \norm{\hvec{j}}}.
\end{equation}
The cosine similarity abstracts away from the magnitude of the vectors, which may be useful in reducing the frequency effect of embeddings \citep{bakarov2018survey} as some embeddings could receive much larger updates than others; see \cite{levy2014linguistic} for a discussion on cosine similarity and vector arithmetic for word embedding intrinsic evaluation. 

Notably, however, we do not evaluate the cosine similarity between a term vector $\hvec{i}$ and a context vector $\hcovec{j}$. Rather, embedding methods typically only release the term vectors as the final output of the model \citep{mikolov2013efficient,mikolov2013distributed} (or, in the case of GloVe, the term and context vectors are averaged together before output \citep{pennington2014glove,levy2015improving}). Therefore, it would be inappropriate to represent the cosine similarity using a notation based on normalizing $\hdot{i}{j}$, as this represents the inner product between a term and a context, not between two term vectors. By moving away a function based on $\hdot{i}{j}$ to a function based on $\hvec{i} \cdot \hvec{j}$ we cannot take advantage of our knowledge of learning objectives in interpreting model output. That is, while we know that the dot product between term and context vectors approximates the PMI --- $\hdot{i}{j} \approx \pmi(i,j)$ --- there are no direct insights garnered from the model as to what $\hvec{i} \cdot \hvec{j}$ approximates. 

In our qualitative analysis in \S\ref{sec:qualitative}, we examine the difference between these two ways to compute word-word similarity. In general, we observe that term-context dot products $\hdot{i}{j}$ will approximately represent the PMI between the two (as expected), while $\hvec{i} \cdot \hvec{j}$ will capture moreso the semantic or syntactic similarity between the two words. This phenomenon has also been observed by \cite{asr2018querying}, who empirically found (without knowledge of the implicit PMI learning objective of these models) that $\hdot{i}{j}$ was better as a measure of asymmetric ``relatedness'' on a dataset specifically designed to only include notions of relatedness \citep{jouravlev2016thematic}, while $\hvec{i} \cdot \hvec{j}$ was better as a measure of ``similarity'' on the Simlex999 dataset \citep{hill2015simlex}. Our work provides a theoretical explanation for \citeauthor{asr2018querying}'s findings, and our qualitative analysis further reinforces them.

\subsection{Tuning to word similarity tasks} \label{sec:tuning}
\begin{table}[t]
    \centering
    \begin{tabular}{l rrrr rrrr}
    \toprule
        \textbf{Model} & B143 & MENd/t & RMT & Rare & SE17 & S999 & W353R/S & Y130\\
    \midrule
        SGNS     & \underline{.453} & .753/.763 & \textbf{.680} & .515 & .656 & .364 & .589/\textbf{.760} & .514 \\
        MF-SGNS  & \textbf{.453} & .730/.754 & .650 & \underline{.559} & \textbf{.684} & \underline{.438} & .565/.751 & .498 \\
        GloVe    & .347 & \underline{.760}/\textbf{.771} & .663 & .509 & .662 & .391 & \underline{.602}/.727 & \underline{.541} \\
        MF-GloVe & .379 & \textbf{.764}/\underline{.768} & .664 & .526 & \underline{.671} & .419 & \textbf{.626}/\underline{.755} & \textbf{.574} \\
        \hilby   & .373 & .733/.747 & \underline{.677} & \textbf{.570} & .668 & \textbf{.453} & .576/.748 & .491 \\
    \bottomrule
    \end{tabular}
    \caption{Performance on word similarity datasets. Best is in \textbf{bold}, second best is \underline{underlined}..}
    \label{tab:similarity-results}
\end{table}
Word similarity tasks are standard ways to perform intrinsic evaluation of word embeddings. However, different datasets focus on different notions of similarity, and in some cases this can lead to quite different performance across tasks. In each dataset, a small set (typically less than two thousand) of pairs of words are obtained. The researcher will then design some kind of crowd-sourced task where typically around 5-10 human judgements per word-pair will be used to measure how ``alike'' (according to the primed notion of likeness) two words are. For example, in Simlex999 \citep{hill2015simlex}, humans are given word pairs and must provide a score between $0$ and $10$ on how ``alike'' the two words are, where, in this dataset, antonyms are treated as completely un-``alike''. Given all of the word pairs, we will use the cosine similarity between word embeddings to represent the model's ``score'' of likeness between the two words. Performance is measured according to the Spearman rank correlation coefficient between the model's scoring of all pairs of words and the gold standard human scoring.

It is necessary to evaluate across a multitude of datasets due to the different notions of likeness used in the different studies. In particular, the notions of \textit{similarity} and \textit{relatedness} have been disentangled for evaluation purposes \citep{agirre2009study,hill2015simlex}. For example, while the two words ``east'' and ``west'' are quite \textit{related}, both being cardinal directions, they are in fact completely \textit{dissimilar} from each other as they are antonyms. Depending on the dataset, they could have a very high or very low human score. Note that, while Simlex999 \citep{hill2015simlex} has been shown to have high correlation with downstream performance on certain tasks \citep{chiu2016intrinsic}, it is still possible to overfit to the task. In particular, \cite{mrkvsic2016counter} introduce counter-fitting (as a specific instance of retro-fitting \citep{faruqui2015retrofitting}) to make antonymous vectors have low cosine-similarity, thus getting state-of-the-art results on Simlex999 at the time. However, this did not correspond to state-of-the-art results on other tasks.

Solving word similarity tasks is not an end in itself, and they are not particularly useful as metrics for downstream performance \citep{chiu2016intrinsic}. We therefore opted to use these tasks to tune the matrix factorization learning rates $\eta$ (Equation~\ref{eq-grad}) for our three models MF-SGNS, MF-GloVe, and \hilby. As we optimized for mean performance across 10 datasets, concerns about overfitting are substantially mitigated. 
We found that model performance was not strongly influenced by the learning rate, so long as we could avoid zero-updates and divergence. 
We tuned the learning rates for each model to: MF-SGNS, $\eta = 0.0005$; MF-GloVe, $\eta=0.005$; \hilby, $\eta=1.0$. For the model hyperparameters we used the same values as the originals (e.g., with respect to context distribution smoothing, empirical weighting function, etc.), as the original designers already tuned and released their models with those parameters. For \hilby, we had to tune $\tau$ along with $\eta$, finding best performance on the word similarity tasks when $\tau=2.0$.

\paragraph{Word similarity results.} In Table~\ref{tab:similarity-results} we present the results for each embedding algorithm across the datasets. Notably, \hilby obtains the best performance on Simlex999 and the Rare-words datasets, suggesting that our model makes high quality embeddings that could correspond to the best performance on the downstream tasks of Named Entity Recognition and NP-chunking \citep{chiu2016intrinsic}, as we neither overfit nor design the model to optimize performance on these specific similarity datasets (unlike \cite{mrkvsic2016counter}). We can also note certain tendencies in model performance that provide evidence that our reimplementations are true to the original algorithms and exhibit similar linear structure in the embedding space. For example, SGNS and MF-SGNS get very similar performance on B143, while GloVe and MF-GloVe both do worse; and this holds in the opposite case for Y130. Additionally, Glove and MF-Glove observe similar performance on the MEN development and test sets, along with WS353R. While there are some differences in performance, these are likely attributable to differences in the implicit minibatch sizes compared to sampling versus our sharding-based MF. Interestingly, the results suggest that MF may create better representations of rare words than sampling; this is not surprising, as every vector receives exactly the same number of updates in MF.

\subsection{Analogical reasoning}
\begin{table}[t]
    \centering
    \begin{tabular}{l rr}
    \toprule
        \textbf{Model} & Google & BATS\\
    \midrule
        SGNS     & \underline{0.763} & 0.312  \\
        MF-SGNS  & \textbf{0.775} & \textbf{0.354}  \\
        GloVe    & 0.739 & 0.310 \\
        MF-GloVe & 0.745 & 0.324 \\
        \hilby   & 0.702 & \underline{0.339}  \\
    \bottomrule
    \end{tabular}
    \caption{Results on analogy datasets; score is proportion of correct analogy completions. Best is in \textbf{bold}, second best is \underline{underlined}.}
    \label{tab:analogy-results}
\end{table}
Analogical reasoning was popularized by \cite{mikolov2013efficient} as a way to evaluate the linear structure captured by the embedding space. The task seeks to use word embeddings to solve the problem of evaluating: $a$ is to $a^*$ as $b$ is to \textit{X} (with the answer being $b^*$). For example, \textit{havanna} is to \textit{cuba} as \textit{moscow} is to \textit{X} (\textit{russia}). Vector embeddings can be used to resolve these analogies with an addition-based method applied on cosine similarities called \textit{3CosAdd}.
However, \cite{levy2014linguistic} proposed the better-performing \textit{3CosMul}, which is more well-motivated than \textit{3CosAdd} based on observations of the linear structure in the embedding space. Our experiments also find that \textit{3CosMul} is consistently better than \textit{3CosAdd}, and we thus only report results from using this retrieval technique. Completing the analogy by using $a, a^*, b$ yields a word $b^*$ as follows:
\begin{equation} \label{eq:3cosmul}
    b^* = \argmax_{b^*} \frac{\cos(b^*, b) \cos(b^*, a^*)}{\cos(b^*, a) + \epsilon}, \qquad \epsilon = 0.001,
\end{equation}
where $\cos(x)$ is shifted to be between $[0, 1]$ \citep{levy2014linguistic}.

\paragraph{Analogical reasoning results.} There has been a considerable amount of work finding that performance on these tasks does not necessarily provide a reliable judgement for embedding quality \citep{faruqui2016problems,linzen2016issues,rogers2017too}. Indeed, from our results in Table~\ref{tab:analogy-results} we observe that \hilby performs markedly worse than the other models on the Google Analogy dataset. However, not only is this uncorrelated with downstream performance (\S\ref{sec:empirical}), it does not even correspond with performance on the other analogy dataset (BATS), where \hilby gets second best performance. Therefore, performance on these tasks should be taken with a grain of salt.

\subsection{Qualitative analysis} \label{sec:qualitative}

In this section we provide a qualitative analysis of the embeddings produced by \hilby. While we focus on only this set of embeddings, we found that the observed qualities are generally consistent across algorithms.

\subsubsection{Dot product or cosine similarity?}
We first examine the impact of using the dot product versus using cosine similarity as a measure of word similarity. We do this because our Simple Embedder framework reveals that each model is trained to produce vectors with term-context dot products that approximate the PMI. At first glance it is not at all intuitively obvious that cosine similarity should be objectively better than the dot product. Qualitatively, we observe the following results when comparing cosine similarity to the dot product on \hilby's embeddings; note that we only use term vectors in this case (as is the standard):

\begin{example*}{Cosine similarity versus dot product similarity (with term vectors)}
\textbf{Cos}: $\argmax_j \cos(\hvec{\textit{hot}}, \hvec{j}) =$ hotter, bubbling, hottest, cool, warm \\
\textbf{Dot}: $\argmax_j \hvec{\textit{hot}}\cdot \hvec{j} =$ hottest, bubbling, hotter, billboard, red-hot\\
\textbf{Cos}: $\argmax_j \cos(\hvec{\textit{cuba}}, \hvec{j}) =$ cuban, cubans, havana, castro, nicaragua \\
\textbf{Dot}: $\argmax_j \hvec{\textit{cuba}}\cdot \hvec{j} =$ cubans, cuban, anti-castro, cuban-americans, havana\\
\textbf{Cos}: $\argmax_j \cos(\hvec{\textit{embedding}}, \hvec{j}) =$ isomorphism, embedded, subset, subsets \\
\textbf{Dot}: $\argmax_j \hvec{\textit{embedding}}\cdot \hvec{j} =$ isomorphism, euclidean, abelian, affine.
\end{example*}
Here we qualitatively observe very little difference between using one or the other with \hilby. This generally holds for SGNS and GloVe as well, but for MF-SGNS and MF-GloVe they can get stranger outlier predictions when using the dot product. Interestingly, for \hilby, performance on Simlex999 actually improves from 0.453 to 0.462 when using the dot product instead of cosine similarity. However, average performance reduces by about 0.022 when using dot products versus cosine similarity for \hilby, which is, perhaps, not as dramatic as a difference as one might suspect. For the other models, however, we found that the performance gap was larger at about 0.05-0.06.

\subsubsection{Term-context or term-term dot products?}
The Simple Embedder framework has allowed us to find that $\hdot{i}{j} \approx \pmi(i,j)$ for \hilby and a PMI variant for the other algorithms. We therefore can anticipate that behavior of term embeddings when multiplied with context embeddings will correspond to their PMIs; i.e., we know what to expect from \textit{term-context} dot products. However, recall that embeddings are either only released with the term vectors \citep{mikolov2013distributed}, or the term and context vectors are averaged together before being used \citep{pennington2014glove}. This means that people do not actually make use of the context vectors separately. Therefore, our Simple Embedder framework does not directly inform us as to what kind of information we can expect from the \textit{term-term} dot products; i.e., $\hvec{i} \cdot \hvec{j} \approx \,?$. Observe the following examples from \hilby:

\begin{example*}{Term versus context vectors with dot product similarity}
\textbf{T-T:} $\argmax_j \hvec{\textit{cat}} \cdot \hvec{j} =$  kittens, cats, kitten, poodle, terrier, dog \\
\textbf{T-C:} $\argmax_j \hvec{\textit{cat}} j \rangle =$ burglar, siamese, scan, scans, tabasco, schr\"odinger\\
\textbf{T-T:} $\argmax_j \hvec{\textit{money}} \cdot \hvec{j} =$ funds, monies, billions, cash, sums \\
\textbf{T-C:} $\argmax_j \hvec{\textit{money}} j \rangle =$ laundering, launder, extort, laundered, funneling \\
\textbf{T-T:} $\argmax_j \hvec{\textit{vector}} \cdot \hvec{j} =$ vectors, tensor, scalar, formula\_34, formula\_10 \\
\textbf{T-C:} $\argmax_j \hvec{\textit{vector}} j \rangle =$ tangent, scalar, subspace, dimensional, non-zero.
\end{example*}
Here, we observe a very substantial difference between $\hdot{i}{j}$ and $\hvec{i} \cdot \hvec{j}$. Namely, $\hdot{i}{j}$ recovers words likely to appear in the context\footnote{Recall that context is defined according to a $w=5$ word window.} of the word $i$: \textit{cat burglar}, \textit{schr\"{o}dinger's cat}, \textit{money laundering}, \textit{tangent vector}, etc\footnote{While ``tabasco'' may seem out of place, ``Tabasco Cat'' is actually a famous American racehorse.}. Meanwhile, $\hvec{i} \cdot \hvec{j}$ corresponds much more strongly to semantic and syntactic relationships. 
Semantically, we observe relationships such as  \textit{cat}-\textit{kittens}, \textit{money}-\textit{funds}, \textit{vector}-\textit{tensor}, etc, that are observed to include both synonyms and antonyms.
Syntactically, we observe relationships like \textit{cat}-\textit{cats}, \textit{money}-\textit{monies}, \textit{vector}-\textit{vectors}. 
We additionally observe some strange relationships, like \textit{vector}-\textit{formula\_34}, which perhaps corresponds to a math formula from Wikipedia that was particularly related to vector spaces, but could be an undesirable recovery\footnote{Another example of an undesirable recovery with term-term embeddings is for the word \textit{take}, which strangely yielded an Atlanta journalist's phone number as the fifth ``most similar'' word: $\argmax_j \hvec{\textit{take}} \cdot \hvec{j} =$ taking, taken, takes, relinquish, 404-526-5456.}. We observe the same behavior concerning these differences between $\hdot{i}{j}$ and $\hvec{i} \cdot \hvec{j}$ when using cosine similarity instead of the dot product, but do not find that it can avoid noisy recoveries.

Overall, these observations may provide insight into how downstream models should utilize word embeddings. In some settings, such as in a text auto-complete system, it would be advantageous to use embeddings to predict surrounding context via term-context dot products $\hdot{i}{j}$. In other settings, such as in an automatic grammar- or meaning-checker, it would be more appropriate to take advantage of the syntactic and semantic information represented in the term-term dot products $\hvec{i} \cdot \hvec{j}$.

\section{Extrinsic Evaluation} \label{sec:empirical}
We perform extrinsic evaluation across a 5 different NLP datasets. We experiment on 8-class news article classification, 2-class sentiment analysis, POS-tag sequence labelling on the WSJ and Brown corpora with different tag syntactic granularity, and supersense-tag sequence labelling on the Semcor 3.0 dataset. For classification, we provide results from three different models with varying levels of depth, in terms of neural network architecture. While this may not definitively determine embedding quality (if anything can), we believe that these tasks exhibit a fair view of performance across different axes of considerations (syntax vs. semantics, sequence labelling vs. classification, deep vs. shallow models etc.) in a way that is relevant for NLP practitioners.
Our experiments are designed in the spirit of ``probing'' the properties of our embeddings \citep{conneau2018what}. As such, we sought to eliminate any factors that might mask the impact of using one set of embeddings versus another. Therefore, every experiment uses \textit{only} the word embeddings as features, and we do not allow any gradient fine-tuning of the embeddings\footnote{During preliminary experimentation we found that gradient fine-tuning did not offer substantial improvements (in some cases it actually degraded validation performance), and it always slowed training time.}.

All presented results are those obtained on the held-out test sets after training the models. Each model was trained 10 times on the same training set with different random seeds for weight initialization in order to examine the model sensitivity that may be induced when using the different embeddings. We therefore present the mean accuracy over the 10 runs along with the standard deviation (for the more sensitive LSTMs). 
For each classification/sequence-labelling model we found that an initial learning rate of $0.001$ worked consistently well across models and datasets when coupled with a learning rate scheduler (other than Logistic Regression). The scheduler worked by evaluating model performance on a previously split-off validation set ($15\%$ of the training set) at the end of every epoch --- if the model could not improve the validation accuracy within 5 epochs, the learning rate was divided by a factor of $10$. The accuracies presented are the test set obtained at the epoch with the best validation accuracy. The bidirectional LSTMs used for both POS-tagging and text classification had two layers with $128$ dimensional hidden units each, structured as described by \cite{huang2015bidirectional}. They were trained with a minibatch size of $16$ for 50 epochs, and had a dropout rate of $0.5$.

\subsection{Text classification}
\begin{table}[t]
    \centering
    \begin{tabular}{l | rrr | rrr}
    \toprule
    \textit{Dataset}
        & \multicolumn{3}{c}{\textit{AGNews}} 
        & \multicolumn{3}{c}{\textit{IMDB sentiment}} \\
        \text{Classifier} & LR & FFNN & BiLSTM & LR & FFNN & BiLSTM \\
    \midrule
    SGNS     &  0.690 & 0.739 & \textbf{0.792} $\pm$ .006 & 
                0.826 & 0.785 &  0.880 $\pm$ 0.010 \\
    MF-SGNS  &  0.697 & \underline{0.752} & 0.785 $\pm$ .009 &
                \textbf{0.829} & \underline{0.807} & \textbf{0.897} $\pm$ 0.003 \\
    GloVe    &  0.663 & 0.747 & 0.775 $\pm$ .013 & 
                0.812 & 0.788 &  0.890 $\pm$ 0.010 \\
    MF-GloVe &  \textbf{0.706} & \textbf{0.759} & 0.783 $\pm$ .008 & 
                \underline{0.828} & \textbf{0.820} & 0.893 $\pm$ 0.008 \\
    \hilby   & \underline{0.704} & \underline{0.752} & \underline{0.790} $\pm$ .004 & 
            0.824 & 0.799 & \underline{0.894} $\pm$ 0.004 \\
    \bottomrule
    \end{tabular}
    \caption{Results for classification; the standard deviation was insignificant for logistic regression (LR) and the feed forward neural network (FFNN) across embedding models. Best is in \textbf{bold}, second best is \underline{underlined}.}
    \label{tab:classification-results}
\end{table}

We performed extrinsic evaluation for classification tasks on two benchmark NLP classification datasets. First, the IMDB movie reviews dataset for sentiment analysis \citep{maas2011learning}, divided into train and test sets of 25,000 samples each. Second, the AGNews\footnote{\url{https://www.di.unipi.it/~gulli/AG_corpus_of_news_articles.html}} news article classification dataset, as divided into 8 approximately 12,000-sample classes (such as \textit{Sports}, \textit{Health}, and \textit{Business}) by \cite{kenyon2018clustering}; for this dataset, we separate out $30\%$ of the samples as the test set. 

On both datasets we experiment with three classification models in increasing order of complexity: logistic regression (LR), a feed-forward neural network with two 128-dimensional hidden layers (FFNN), and a two-layer bidirectional LSTM that concatenates the forward and backward hidden states (128-dimensional) before prediction. For LR and FFNN, we use \textit{max}-pooling of the word embeddings in a sequence as the input features\footnote{We found \textit{mean}-pooling to be consistently worse.}, motivated by the discussion of \cite{shen2018baseline} who argue for testing such models to accompany analysis of the performance of deeper models. 
Both LR and the FFNN used a minibatch size of 16, trained for 250 epochs, and the FFNN had ReLU activations, a dropout rate of $0.5$, and two $128$-dimensional hidden layers.

\paragraph{Text classification results.}
Table~\ref{tab:classification-results} presents the results for three classification models applied with only the word embeddings as feature input for two classification problems. We first observe that our MF reimplementations for GloVe and SGNS almost always do better than the original models, despite all being trained on the same dataset. We also observe that certain models seemed to prefer some embeddings over others. While MF-Glove tends to get the best results when used in Logistic Regression and an FFNN, the BiLSTM seemed to prefer \hilby embeddings consistently (second-best for \textit{both}), and traded off between SGNS (\textit{news}) and MF-SGNS vectors (\textit{sentiment}) for first place. We also note that, in the case of sentiment analysis, despite being provided the same feature inputs, using an FFNN can actually damage performance. However, it is likely that more advanced forms of embedding pooling would improve an FFNN's results, given the much higher performance obtained by the BiLSTM. Lastly, we observe that \hilby embeddings were by far the most resistant to the random initialization of the weight vectors in the BiLSTM models, given the low accuracy variance of $0.004$. These results all suggest that \hilby embeddings are highly reliable across a wide range of classification tasks and models.

\subsection{Sequence labelling}
Our final set of experiments were sequence labelling tasks for three different datasets and tagsets: 12-label POS-tagging on the Brown corpus; 44-label POS-tagging on the WSJ PTB corpus; and 83-label semantic supersense tagging on the Semcor3.0 corpus. In each task, we used a 2-layer bidirectional LSTM (BiLSTM) with forward and backward hidden representations concatenated before label prediction.

\paragraph{Part-of-speech tagging.}
Part-of-speech tagging is a well-studied task in NLP where the goal is to predict the syntactic labels (such as \textit{nouns} and \textit{verbs}) for each token in a sequence of text. We experiment with two corpora with different levels of label granularity. At a finer granularity, we use the Wall Street Journal corpus (WSJ) \citep{marcus1993building} with its standard 44-label tagset, and at a lower granularity we use the Brown corpus (as distributed by NLTK\footnote{\url{https://www.nltk.org/book/ch02.html}}) mapped to the 12-label ``universal'' tagset of \cite{petrov2012universal}.
For \textit{WSJ}, we used the standard train-test split with sections 22, 23, and 24 as the test set and the rest as the training set, corresponding to 1.1 million tokens for training (44,000 sequences) and 138,000 for testing (5,500 sequences). 
For the \textit{Brown corpus} we randomly divided 30$\%$ of the sequences as the test set, with the other 70\% as the training set. This corresponded to 800,000 tokens for training (40,000 sequences) and 350,000 for testing (17,000 sequences).

\paragraph{Supersense tagging.}
Supersense tagging (SST) is form of semantic labelling of text sequence, but is a much less well-studied sequence labelling problem than POS-tagging. Introduced by \cite{ciaramita2006broad}, SST includes 25 semantic labels for nouns (including basic semantic concepts such as  \textit{animal} and \textit{artifact}, and more abstract ones such as \textit{attribute} and \textit{relation}) and 16 labels for verbs (e.g., \textit{communication} for the verb ``asking'' and \textit{motion} for the verb ``flying''). \citeauthor{ciaramita2006broad} introduced it as a coarse grained version of word sense disambiguation by defining a small set of possible labels in order to be solvable with standard sequential machine learning models of the time (such as HMMs). These labels are obtained from \textit{Wordnet}'s lexicographic classes, so any dataset labelled for word sense disambiguation induces a dataset with supersense labels by following the word sense hierarchy up to the top lexicographic classes. As such, we use the latest version (3.0) of the Semcor corpus \citep{miller1993semantic} and divide it into training and test sets with 305,000 and 129,000 tokens each (14,000 and 6,000 sequences), respectively.

SST uses standard B-I-O tagging to deal with multi-word expressions. It therefore encompasses Named Entity Recognition due to the fact that named entities comprise a subset of the tags (such as \textit{group} and \textit{person}) used in SST \citep{ciaramita2006broad}. 
In the example below, we see that the named entity ``Empire State Building'' is represented by the three token level supersense tags: first \textit{B.n.artifact} and then two \textit{I.n.artifact} tags (abbreviated in the example with just \textit{I} for ease of exposition):
\begin{example*}{Supersense tagging}
    \textbf{The}$_O$ \textbf{Empire}$_{B.n.artifact}$ \textbf{State}$_{I}$ \textbf{Building}$_{I}$ \textbf{may}$_O$ \textbf{be}$_O$ \textbf{destroyed}$_{B.v.change}$ \textbf{by}$_O$ \textbf{terrorists}$_{B.n.person}$ \textbf{on}$_O$ \textbf{Thursday}$_{B.n.time}$ \textbf{.}$_O$
\end{example*}

Thus, while there are 25+16 $=$ 41 supersenses, each one must also have a \textit{B[eginning]} and \textit{I[ntermediate]} tag, plus a label for items that do not have any tag, \textit{O[utside]}. Therefore, this is a sequence labelling problem with exactly 83 possible labels for each token. 
Note that about $66\%$ of the tokens are labelled with the \textit{O} tag, so it is standard to report results using the micro-F-score without the \textit{O}'s score due to the skew of label distribution \citep{alonso2017multitask,changpinyo2018multi}.

\paragraph{Sequence labelling results.}
\begin{table}[t]
    \centering
    \begin{tabular}{lccc}
    \toprule
        \textit{Dataset} & \textit{Semcor SST} & \textit{WSJ POS} & \textit{Brown POS}\\
        (metric) & (micro-F1) & (accuracy) & (accuracy)\\
    \midrule
        Most-frequent baseline & 0.6126 & 0.8905 & 0.9349 \\
        \hline
        SGNS     & 0.6638 $\pm$ .0014 & 0.9615 $\pm$ .0004 & 0.9762 $\pm$ .0002 \\
        MF-SGNS  & \textbf{0.6696} $\pm$ .0027 & \textbf{0.9621} $\pm$ .0005 & \textbf{0.9771} $\pm$ .0002 \\
        GLoVe    & 0.6550 $\pm$ .0033 & 0.9609 $\pm$ .0004 & 0.9750 $\pm$ .0003 \\
        MF-GLoVe & 0.6655 $\pm$ .0026 & 0.9613 $\pm$ .0003 & 0.9759 $\pm$ .0003 \\
        \hilby   & \underline{0.6660} $\pm$ .0020 & \underline{0.9616} $\pm$ .0004 & \underline{0.9767} $\pm$ .0003 \\
    \bottomrule
    \end{tabular}
    \caption{Sequence labelling performance using a standard bidirectional LSTM, accuracy plus standard deviation over 10 runs with different random seeds for initialization. Best is in \textbf{bold}, second best is \underline{underlined}.}
    \label{tab:seqlab-results}
\end{table}

To accompany our results in Table~\ref{tab:seqlab-results}, we include results from a trivial \textit{Most-frequent tag baseline}. This baseline simply returns that the tag of a token is the tag that most frequently occurs for that token within the training set. We include this simply as a ``sanity check'' to verify that our embeddings and model are generalizing well. Indeed, in supersense tagging (SST) it is standard to include results from the most-frequent-\textit{supersense} baseline, since it is inspired from the tradition of word sense disambiguation \citep{ciaramita2006broad}, which uses the most-frequent-\textit{sense} as a baseline. 
We observe that our models generalize substantially over the baseline. 
Note that we do not obtain very strong results on WSJ in comparison with some conditional random field models, which can get an accuracy of 97.36\% \citep{yao2014recurrent}. This is expected, however, as we do not include any hand engineered character- or context-based features. Such features are necessary to incorporate into a standard BiLSTM in order to obtain performance above 97.30\% \citep{huang2015bidirectional}. Yet, using only word embeddings as features\footnote{In particular, the ``Senna'' word embeddings of \cite{collobert2011natural} that were trained for over two months on Wikipedia (\url{http://ronan.collobert.com/senna/}).}, \citeauthor{huang2015bidirectional} report results of 96.04\% with their BiLSTM, which suggests that our embeddings are stronger and that our deep model functions well, given that MF-SGNS gets a score of 96.21\% and \hilby gets second place with 96.16\%\footnote{Note that this seemingly small percent difference  (96.21 to 96.04) corresponds to over 230 tokens being labelled correctly on the test set.}. Overall, we observe very consistent results across the tasks. MF-SGNS's results are the best (but at a higher variance, for SST and WSJ-POS) while \hilby gets second best results with lower variance. 

These results also point to the difficultly of supersense tagging.  
There are many studies on multi-task learning for sequence labelling that include results for supersense tagging \citep{sogaard2016deep,alonso2017multitask,changpinyo2018multi}. Additionally, deep models have been specialized for problems such as word sense disambiguation \citep{raganato2017neural}. However, aside from the very interesting study on building supersense embeddings by \cite{flekova2016supersense}, there is no work to our knowledge on designing specialized embedding features as input to deep models for SST; we leave it to future work to improve upon these results.

%%% ------------------------------------ %%%
%%% ------------------------------------ %%%
%%% ------ Now for the Conclusion ------ %%%
%%% ------------------------------------ %%%
%%% ------------------------------------ %%%
\chapter{Conclusion}
We have proposed the Simple Embedder framework to generalize existing word embedding algorithms --- including Word2vec (SGNS) \citep{mikolov2013distributed}, GloVe \citep{pennington2014glove}, and other algorithms --- within the \textit{generalized low rank model} family of matrix factorization algorithms \citep{udell2016generalized}.
Our wide variety of empirical results provide reliable evidence that our theoretical proofs are sound, due to the consistency in results between our matrix factorization reimplementations and the original SGNS and GloVe algorithms. 
% Our theoretical and empirical results reveal that our generalizing framework is correct, 
Indeed, we find that the matrix factorization reimplementations tend to offer slightly better performance than the originals, especially in terms of better representations of rare words.
Our primary theoretical contribution is that, once cast as Simple Embedders, comparison of these models (in terms of their gradients and objectives) reveals that these successful embedders all resemble, analytically and empirically, a straightforward maximum likelihood estimate (MLE) of inner-product parametrized PMI. This MLE induces our proposed novel word embedding algorithm, \hilby.

Our algorithm is theoretically the most parsimonious of the algorithms within the Simple Embedder framework. 
\hilby substantially reduces the number of hyperparameters (unlike \cite{arora2016latent}) that are required for the other algorithms to work \citep{levy2015improving}. 
Moreover, \hilby is consistently at the high end of the characteristic level of performance of these models across 12 intrinsic and 5 extrinsic evaluation tasks. In particular, \hilby consistently observes second-best performance on every extrinsic evaluation (news-article classification, sentiment analysis, POS tagging on the WSJ and Brown corpora, and supersense tagging), while the first-best model depends varying on the task. Moreover, \hilby consistently observes the least variance in results with respect to the random initialization of the weights in bidirectional LSTMs. These empirical results suggest that \hilby is a highly consistent word embedding algorithm that can be reliably integrated into existing NLP systems to obtain high-quality results.

\section{Directions for Future Work}
The Simple Embedder framework released with this work facilitates rapid training of different types of word embeddings and experimentation over a large range of hyperparameters. For our example, after parsing the corpus to obtain $N_{ij}$ statistics, any algorithm can be run on top of them; MF-SGNS produced the 50,000-vocabulary set of embeddings used in our experiments in less than one hour on a single GPU, and we later found that \hilby can use a larger learning rate of $\eta=8.75$ to train at a similarly rapid speed. 

It is highly probable that different algorithms, different hyper-parameters, and especially different definitions of \textit{context} \citep{li2017investigating} will produce embeddings with different expressive qualities than any one set of embeddings could capture on their own. This therefore suggests that it would be worthwhile to produce a large set of differently-trained embeddings using our framework, making an \textit{an ensemble of word embeddings}. Indeed, our embeddings should be used within meta-embeddings \citep{yin2016learning} and dynamic meta-embeddings \citep{kiela2018dynamic} to obtain state-of-the-art results on NLP tasks.

In conclusion, our theoretical and practical findings in this work advance the collective understanding of word embedding technology. Our findings for these relatively simple algorithms can provide a foundation for investigations into the properties of more complex deep contextualized models \citep{peters2018deep,devlin2018bert}. We believe that such investigations of more advanced models require --- as a necessary prerequisite --- a solid theoretical understanding of the simpler word embedding algorithms that preceded them, and we hope that this work provides inspiration for future researchers in such pursuits.

\newpage

%%%%%%%%%%%%%%%%%%%
%
% BIBLIOGRAPHY
%
%%%%%%%%%%%%%%%%%%%

%\chapter*{\rm\bfseries Bibliography}
\chaptermark{{\rm\bfseries Bibliography}}
\addcontentsline{toc}{chapter}{Bibliography}

\bibliographystyle{apalike}
\bibliography{Thesis}

%\addcontentsline{toc}{chapter}{\numberline{}\rm\bfseries{Bibliography}}

\begin{onehalfspacing}

\chapter*{{\huge\rm\bfseries{Acronyms}}}

\addcontentsline{toc}{chapter}{Acronyms}
\scriptsize
\begin{table}[!h]
\vspace{-2.0cm}
\hspace{1.0cm}
\begin{tabular}{l l}

\textsc{CNN}   &  Convolutional neural network \\
\textsc{CRF}   &  Conditional random field \\
\textsc{FFNN}  &  Feed-forward neural network \\
\textsc{GPU}   &  Graphics processing unit \\
\textsc{GB}    &  Gigabytes (of memory)\\
\textsc{HAL}   &  Hyperspace Analogue to Language \\
\textsc{HMM}   &  Hidden Markov model \\
\textsc{LSTM}  &  Long short-term memory neural network \\
\text{BiLSTM}  &  Bidirectional LSTM \\
\textsc{LR}    &  Logistic regression \\
\textsc{MF}    &  Matrix mactorization \\ 
\textsc{MLE}   &  Maximum-likelihood estimation \\
\textsc{NLP}   &  Natural Language Processing \\
\textsc{NLTK}  &  Natural language toolkit (\url{nltk.org})\\
\textsc{OOV}   &  Out of vocabulary \\
\textsc{PCA}   &  Principal component analysis \\
\textsc{PMI}   &  Pointwise mutual information \\
\textsc{POS}   &  Part-of-speech \\
\text{ReLU}    &  Rectified linear unit activation function; $ReLU(x)=\max(0, x)$\\
\text{RNN}     &  Recurrent neural network \\
\textsc{SE}    &  Simple embedder \\
\textsc{SGNS}  &  Skip-Gram with Negative Sampling \\
\textsc{SST}   &  Supersense tagging \\
\textsc{SVD}   &  Singular value decomposition \\
\textsc{WSJ}   &  Wall Street Journal corpus \\

\end{tabular}
\end{table}

\chapter*{{\huge\rm\bfseries{Notation}}}

\addcontentsline{toc}{chapter}{Notation}
\scriptsize
\begin{table}[!h]
\vspace{-2.0cm}
% \hspace{1.0cm}
    \centering
    \begin{tabular}{ll}
        \textit{Symbol} & \textit{Meaning} \\ \midrule
        $\corpus$ & a combinatorial structure of a language in use, e.g., the corpus of text \\
        $\vocab$ & a set representing the entire vocabulary used by a SE applied on $\corpus$\\
        $\vocab_T$ & a subset of $\vocab$, the \textit{term} vocabulary with number of elements $|\vocab_T|$\\
        $\vocab_C$ & a subset of $\vocab$, the \textit{context} vocabulary with number of elements $|\vocab_C|$\\
        $d$ & the desired dimensionality of the embeddings to be produced by a SE\\
        $i,j$ & $i$ is used to index the \textit{terms} and $j$ is used to index the \textit{contexts}  \\
        $\hvec{i}$ & the term embedding vector for term $i$ (a row vector in $\R^d$) \\
        $\hcovec{j}$ & the context embedding vector for context $j$ (a column vector in $\R^d$) \\
        $\hdot{i}{j}$ & the inner product between $\hvec{i}$ and $\hcovec{j}$ \\
        $\vectors$ & the matrix of all term vectors $\hvec{i}$ in $\R^{|V_T| \times d}$\\
        $\covectors$ & the matrix of all context vectors $\hcovec{j}$ in $\R^{d \times |V_C|}$\\
        $\hat{\mathbf{M}}$ & the product $\vectors \covectors$ with $\hat{\mathbf{M}}_{ij} = \hdot{i}{j}$\\
        $\mathcal{L}$ & a generic global loss function for a SE\\
        $\Delta$ & matrix in $\R^{|\vocab_T|\times |\vocab_C|}$ of partial derivatives of $\mathcal{L}$ with respect to $\hat{\mathbf{M}}$ \\
        $f_{ij}$ & item of the summation in $\mathcal{L}$ corresponding to term-context pair $i,j$ \\
        $\phi_{ij}$ & measure of association between $i,j$ such that $\hdot{i}{j}=\phi_{ij}$ minimizes $f_{ij}$ \\ 
        $\mathbf{M}$ & matrix in $\R^{|\vocab_T|\times |\vocab_C|}$ filled with $\phi_{ij}$ in each element\\
        $w$ & context window size (typically $w=5)$\\
        $\Omega$ & the set of every $i,j$ cooccurrence (according to $w$) in the corpus $\corpus$\\
        $N$ & total number of $i,j$ cooccurrences, $N=|\Omega|$\\
        $N_{ij}$ & number of times term-context pair $i,j$ occurs in the corpus\\
        $N_i$ & $\sum_{j} N_{ij}$, number of times term $i$ occurs in $\Omega$\\
        $N_j$ & $\sum_{i} N_{ij}$, number of times context $j$ occurs in $\Omega$\\
        $p_i$ & $N_i / N$, unigram (marginal) probability of term $i$\\
        $p_j$ & $N_j / N$, unigram (marginal) probability of context $j$\\
        $p_{ij}$ & $N_{ij} / N$, cooccurrence (joint) probability of $i,j$\\
        $p^I_{ij}$ & $p_i p_j$, probability of $i,j$ cooccurrence \textit{under independence}\\
        $\hat{p}_{ij}$ & $(N_i N_j / N^2) \exp{\hdot{i}{j}}$, a SE's approximation of the joint probability\\
        $k$ & number of negative samples (for SGNS) \\
        $\log$ & the natural logarithm (base $e$) \\
        $\cos$ & the cosine similarity function between vectors (Equation~\ref{eq:cosinesim}) \\
        $\sigma$ & the logistic sigmoid function $\sigma(x) = 1 / (1 + \exp(-x))$\\
    % \bottomrule
    \end{tabular}
    \caption{Notation used in this work; SE stands for \textit{simple embedder}.}
    \label{tab:notation}
\end{table}

\end{onehalfspacing}

\normalsize
\printindex

% \appendix
% \input{content/ConsADExamples.tex}

\end{document}